%% file: main.tex
\documentclass[a4paper]{svproc}
\usepackage{wrapfig}

\usepackage[T1]{fontenc}
\def\doi#1{\href{https://doi.org/\detokenize{#1}}{\url{https://doi.org/\detokenize{#1}}}}
\usepackage{graphicx}
\usepackage{subcaption}
\usepackage{hyperref}

\usepackage{amsmath,amsthm,amssymb,bbm} 
\usepackage{breqn}  
\usepackage{algorithmicx} %
\usepackage{algorithm}%
\usepackage[noend]{algpseudocode}%

\usepackage{enumitem} %
\usepackage{xcolor}

\usepackage{multirow}

\usepackage{capt-of}
\algrenewcommand\algorithmicindent{1em}%

\usepackage{listings}

\input{macros}
\algrenewcommand\algorithmicrequire{\textbf{Input:}}
\algrenewcommand\algorithmicensure{\textbf{Output:}}

\usepackage[font=small,labelfont=bf]{caption}

\makeatletter
\def\thanks#1{\protected@xdef\@thanks{\@thanks
        \protect\footnotetext{#1}}}
\makeatother

\begin{document}
\title{Robust-RRT: Probabilistically-Complete Motion Planning for Uncertain Nonlinear Systems\thanks{The NASA University Leadership initiative (grant \#80NSSC20M0163), the National Science Foundation, Cyber-Physical Systems (CPS) program (award 1931815), and NVIDIA provided funds to assist the authors with their research. This article solely reflects the opinions and conclusions of its authors. %
}}
\titlerunning{Probabilistically-Complete Motion Planning for Uncertain Nonlinear Systems}
\author{Albert Wu\inst{1}%
\and
Thomas Lew\inst{1}%\inst{2}%
\and
Kiril Solovey\inst{2}%\inst{3}%
\and\\
Edward Schmerling \inst{1}%\inst{2}
\and
Marco Pavone\inst{1}%\inst{2}
}
\authorrunning{Wu et al.}
\vspace{-2mm}
%
%
%
% \institute{Department of Computer Science, Stanford University, Stanford, %
% USA\\ \email{amhwu@stanford.edu} \and
% Department of Aeronautics and Astronautics, Stanford University, Stanford, %
% USA\\ \email{\{thomas.lew,schmrlng,pavone\}@stanford.edu} \and
% Faculty of Electrical and Computer Engineering, Technion, Haifa, Israel\\
% \email{kirilsol@technion.ac.il}
% }
\institute{Stanford University, Stanford, %
USA\\ \email{\{amhwu,thomas.lew,schmrlng,pavone\}@stanford.edu} \and
Technion, Haifa, Israel\\
\email{kirilsol@technion.ac.il}
}
\maketitle              %
\vspace{-2mm}
\begin{abstract}
Robust motion planning entails computing a global motion plan that is safe under all possible uncertainty realizations, be it in the system dynamics, the robot's initial position, or with respect to external disturbances. Current approaches for robust motion planning either lack theoretical guarantees, or make restrictive assumptions on the system dynamics and uncertainty distributions.
In this paper, we address these limitations by proposing the robust rapidly-exploring random-tree (\texttt{Robust-RRT}) algorithm, which integrates forward reachability analysis directly into sampling-based control trajectory synthesis. %
We prove that \texttt{Robust-RRT} is probabilistically complete (PC) for nonlinear Lipschitz continuous dynamical systems with bounded uncertainty. In other words, \texttt{Robust-RRT} eventually finds a robust motion plan that is feasible under all possible uncertainty realizations assuming such a plan exists. 
Our analysis applies even to unstable systems that admit only short-horizon feasible plans; this is because we explicitly consider the time evolution of reachable sets along control trajectories. 
% Thanks to the explicit consideration of time dependency in our analysis, PC applies to unstabilizable systems. 
To the best of our knowledge, this is the most general PC proof for robust sampling-based motion planning, in terms of the types of uncertainties and dynamical systems it can handle.
Considering that an exact computation of reachable sets can be computationally expensive for some dynamical systems, we incorporate sampling-based reachability analysis into \texttt{Robust-RRT} and demonstrate our robust planner on nonlinear, underactuated, and hybrid systems.
%
%

% \keywords{robust motion planning, planning under uncertainty, reachability analysis, sampling-based motion planning.}

\end{abstract}
%
% Reset the footnote numbering to exclude \institute
\setcounter{footnote}{0} 

% \newpage
%
%
%
%
%
%
%
%
%
%
%
%
%
%
%
%
%
%
%
%
\vspace{-5mm}
\section{Introduction}
\label{sec:introduction}
\vspace{-2mm}
Motion planning algorithms are an integral part of the robotic autonomy stack and typically entail computing a \textit{feasible} plan, which satisfies the system dynamics, respects all constraints, avoids obstacles, and reaches the goal.
In safety-critical applications, it is important to ensure feasibility holds under uncertainty, such as external disturbances and uncertain model parameters.

This paper addresses the task of computing \textit{robust motion plans}, i.e., motion plans that can be safely executed %
under all possible uncertainty realizations belonging to a bounded set.
In particular, we seek the following desiderata in robust motion planning algorithms:
\begin{itemize}[leftmargin=4mm]\setlength\itemsep{1mm}
    \item \textit{Generality: } 
    The algorithm should be applicable to a broad class of complex systems (e.g., nonlinear, underactuated, and hybrid systems)
    with both \textit{epistemic} uncertainty (with respect to, e.g., model parameters)  and \textit{aleatoric} uncertainty (with respect to, e.g., external disturbances) %
    % Specifically, in this paper we consider systems that may not even by stabilizable (e.g., modeling transient loss of control while steering between stabilizable regions).
    that may only admit short-horizon feasible plans (e.g., systems that are not stabilizable everywhere or induce transient loss of control while steering between invariant sets).
    \item \textit{Completeness: } The algorithm should return a robust motion plan if there exists one. In the context of randomized motion planners, it should be \textit{probabilistically complete} (PC), i.e., it should eventually discover the robust solution with probability one. %
\end{itemize}

Existing robust motion planners in literature have yet to satisfy these desiderata. Some planners strive for theoretically-sound robustness against external disturbances. However, they are typically quite conservative (e.g., by ``padding'' constraints globally, %
a homotopy class that is robustly feasible for the original problem could become artificially infeasible for the algorithm), cannot handle epistemic uncertainty (e.g., parametric uncertainty induces time correlations along state trajectories, which are challenging to account for), and are restricted to specific systems (e.g., linear dynamics). Other planners pursue generality across system types and uncertainty sources at the cost of theoretical justifications.
Achieving general and complete robust planning remains an open challenge.
\\[1mm]
\textbf{Contributions.} 
We present a general-purpose robust motion planning framework, analyze its theoretical properties, and address practical considerations.%
\begin{enumerate}[leftmargin=5mm]\setlength\itemsep{1mm}
    \item 
    \textbf{\robustrrt}: We propose a motion planner that integrates forward reachability analysis with sampling-based control trajectory synthesis, %
    and accounts for both aleatoric and epistemic uncertainty. %
    By directly analyzing reachable sets, \robustrrt does not introduce additional conservatism. %
    \item \textbf{Probabilistic completeness}: We prove that \robustrrt is PC. By constructing a \textit{growing funnel} in the space of reachable sets, our analysis only assumes bounded Lipschitz continuous dynamics and bounded uncertain parameters and disturbances. 
    We also discuss generalizing our arguments to RRTs with conservative reachable set estimates and other types of sets.
    \item \textbf{Practical solutions}: 
    We provide concrete solutions to the practical challenges of reachable set-based planning. \robustrrt maintains a single representative state of each reachable set for tree expansion, which allows node sampling to be implemented with standard point-based nearest neighbor algorithms. This avoids the costly ``nearest sets'' computation and retains Voronoi biases for rapid exploration. 
    Additionally, as exact computation of reachable sets can be expensive, we  %
    propose the \randuprrt extension. By 
    approximating reachable sets with forward dynamics rollouts, \randuprrt is applicable to a wide range of systems including black-box physics simulators with no requirement on the uncertainty structure.

\end{enumerate}
\begin{figure}[!ht]
    \centering
    \vspace{-3mm}
    \includegraphics[width=0.7\textwidth]{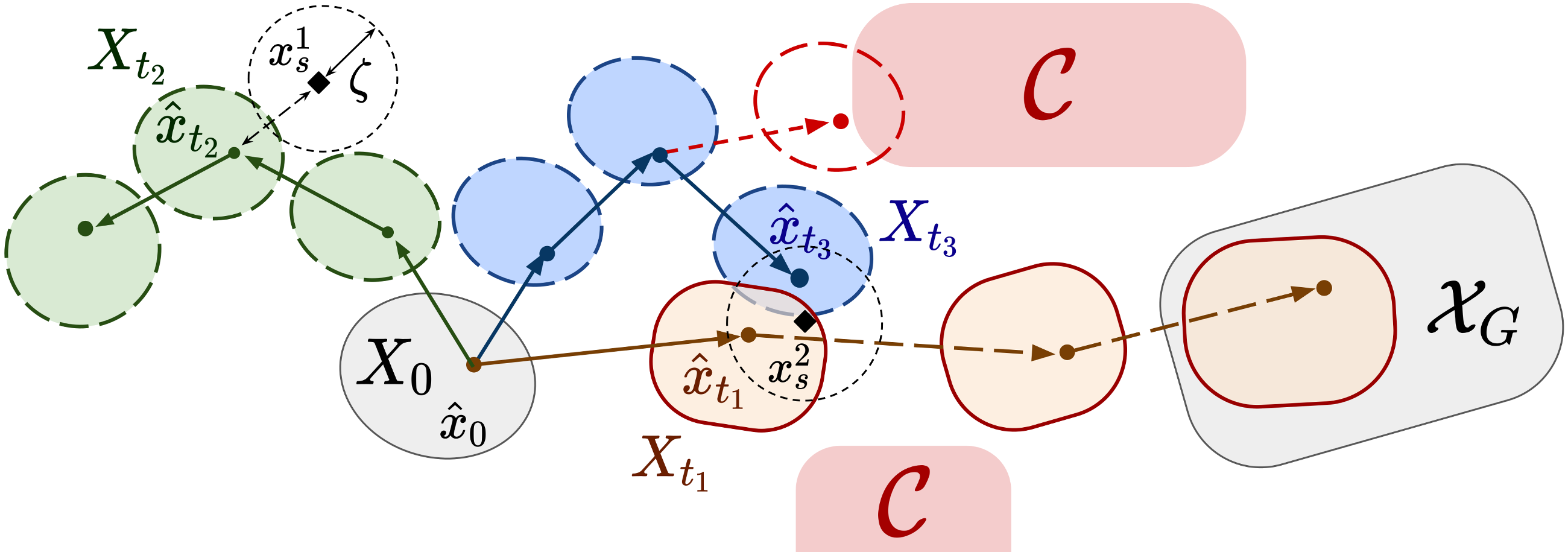}
      \caption{\robustrrt constructs a tree of reachable sets $X_{t_i}$ to compute a robustly feasible trajectory to the goal region $\X_G$ that avoids all unsafe regions $\Obs$. 
      \robustrrt also maintains a tree of nominal states $\hat{x}_{t_i}$ to guide node selection for expanding the reachable set tree, see Algorithm \ref{alg:sample_node}. 
      Specifically, we first sample a state $x_{s}$ from the state space. If the distance from $x_{s}$ to all nominal states is greater than a constant $\zeta$, 
      tree extension is performed from the reachable set that corresponds to the nearest nominal state (e.g., after sampling $x_{s1}$,  the tree is extended from $X_{t_2}$). 
      Otherwise, we randomly choose a nominal state among those within a distance $\zeta$ from $x_s$ 
      (e.g., after sampling $x_{s2}$, one extends from $X_{t_1}$ or $X_{t_3}$). 
      This node selection strategy encourages rapid exploration and circumvents the need to compute distances to reachable sets.}
    \label{fig:node_sampling}
    % \vspace{-6mm}
\end{figure}
We validate \randuprrt on robust motion planning tasks on nonlinear, underactuated, and hybrid systems. Our examples cover uncertain system parameters, external disturbances, unstabilizable systems, and uncertain guard surface locations.%
\\[2mm]
\textbf{Organization.} 
We review related work in Section \ref{sec:related_work} and define the robust motion planning problem of interest in Section \ref{sec:problem_formulation}. We present \robustrrt in Section~\ref{sec:robust_rrt_algorithm} and prove its probabilistic completeness in Section~\ref{sec:pc_proof}. We present practical considerations in Section \ref{sec:practical_implementation} and empirical results in Section \ref{sec:experiments}.%

\vspace{-3mm}
\section{Related Work}
\label{sec:related_work}
\vspace{-2mm}

\textbf{Rapidly-Exploring Random Trees (\rrt)} is one of the most popular samp-ling-based motion planning algorithms in literature \cite{lavalle_2001}. The simplicity and efficiency of \rrt have given rise to variations for addressing a broad range of applications, including planning kinodynamically-feasible plans for nonholonomic and hybrid systems \cite{jaillet2011eg,shkolnik2009reachability,wu2020r3t}. 
The theoretical guarantees of \rrt, such as PC, have been extensively studied in the literature on motion planning for deterministic systems. 
We refer the reader to \cite{chow1940systeme,kleinbort2018,li2016asymptotic} for a thorough discussion.
\\[1mm]
\textbf{Motion planning under uncertainty} extends planning problems to account for uncertainty. These include \textit{aleatoric} (e.g., external additive disturbances that are independent across time) and \textit{epistemic} uncertainty (e.g., parametric uncertainty intrinsic to the system). 
When probability distributions over the uncertain quantities are available, \textit{chance-constrained} motion planning may be leveraged to compute plans that satisfy probabilistic constraints \cite{Aghamohammadi2014FIRMSF,lathrop2021distributionally,lindemann2021robust,liu2014incremental,Luders2011ProbabilisticFF,luders2010chance,melchior2007particle,summers2018,wang2020moment}. 
However, in many applications, only the bounds on the uncertain quantities are available (e.g., when constructing confidence sets for the parameters of a model \cite{abbasi2011improved}), and
one should guarantee safe operation for all uncertainty realizations within these bounds. %
This task is referred to as \textit{robust motion planning}.
\\[1mm]
\textbf{Robust motion planning} algorithms compute motion plans that ensure constraint satisfaction for all possible uncertain quantities within known bounded sets. A robust plan may take the form of a state or nominal control trajectory and may be tracked with a pre-specified feedback controller during execution. The controller may be obtained from control synthesis techniques such as sum-of-squares (SOS) programming \cite{SinghChenEtAl2018,tedrake2010lqr} or Hamilton-Jacobi analysis \cite{bansal2017hamilton,herbert2017fastrack}.
A standard approach to robust motion planning consists of computing offline error bounds (e.g., robust invariant sets), which are subsequently used as robustness margin for planning at runtime \cite{danielson2020robust,herbert2017fastrack,majumdar2016,SinghChenEtAl2018}. As these bounds are independent of the time and state, this procedure is typically conservative and does not allow for PC guarantees. 
In addition, such methods would not return feasible solutions for systems that are not stabilizable, for which no global invariant set exists (see Section \ref{sec:experiments}). 
The approach proposed in~\cite{luders2014optimizing} accounts for uncertainty accumulation over time, but is restricted to linear systems with additive disturbances. 
Importantly, all aforementioned approaches do not account for epistemic uncertainty (e.g., uncertain model parameters). Such uncertainty introduces time correlations along the state trajectory that are challenging to account for. 

We address the aforementioned limitations by tightly integrating forward reachability analysis with RRT. Reachability analysis characterizes the set of states a system may reach at any given time\footnote{Our use of reachability analysis (i.e., the set of states that can be reached for a fixed control trajectory and all possible uncertain parameters) contrasts with reachability analysis in deterministic settings \cite{jaillet2011eg,shkolnik2009reachability,wu2020r3t}, and Hamilton-Jacobi-Isaacs analysis \cite{bansal2017hamilton,kalise2020robust}, which studies which %
states can be reached for all possible control inputs.}. By constructing an RRT with reachable sets as nodes, our approach accounts for the time evolution of uncertainty over trajectories. Constructing a tree of sets to account for aleatoric uncertainty was previously proposed in \cite{panchea2017extended,pepy2009reliable}. However, these approaches perform nearest-neighbor search by computing the Hausdorff distance between set nodes, which is computationally expensive for general set representations. In practice, the above approaches %
compute rectangular approximations of the reachable sets, which eases the computational burden but introduces conservatism. We circumvent this limitation by constructing a state-based tree for \rrt expansion alongside a reachable set-based tree for planning. Importantly, we prove that our proposed approach is PC.
Finally, we leverage recent advances in sampling-based reachability analysis and propose a practical implementation: we interface \rrt with \randup \cite{lew2020samplingbased,LewJansonEtAl2022}, which approximates the true reachable sets with the convex hull of forward dynamics rollouts. \randup enjoys asymptotic and finite-sample accuracy guarantees and enables a simple implementation that can tackle complex systems, is agnostic to the choice of feedback controller used for tracking at runtime, and accounts for both aleatoric and epistemic uncertainty. 

\vspace{-3mm}
\section{Problem Formulation}
\label{sec:problem_formulation}
\vspace{-2mm}

We consider the problem of planning trajectories for uncertain nonlinear dynamical systems navigating in cluttered environments. 
Specifically, we consider uncertain dynamics
\begin{equation}
    \label{eqn:system_dynamics}
    \dot{x}_t = {}^{\theta}f(x_t,u_t,w_t),
    \quad t\geq 0,
\end{equation}
where $x_t\in\X\subset\R^n$ denotes the state of the system at time $t\geq 0$, 
$u_t\in\U\subset\R^m$ denotes control inputs, 
$\theta\in\Theta\subset\R^p$ 
denotes uncertain parameters, 
$w$ represents bounded external disturbances as a stochastic process with bounded total variation such that $w_t\in\W\subset\R^d$ for all times $t\geq 0$, and 
${}^{\theta}f:\R^n\times\U\times\W\to\R^n$ is continuous for all $\theta\in\Theta$.  
We assume that the sets of feasible controls $\U$, 
parametric uncertainty $\Theta$, and 
disturbances $\W$ are compact: these sets represent prior knowledge about the system to bound the level of uncertainty. The probability distributions of $\theta\in\Theta$ and $w_t\in\W$ are unknown, which motivates a robust problem formulation. %

Given a compact set of uncertain initial states $X_0\subset\X$, 
a goal region $\X_G\subset\R^n$ and
a set of obstacles $\Obs\subset\R^n$,  
the problem consists of computing a piecewise-constant open-loop control trajectory $u:[0,t_n]\to \U$ such that 
the system reaches the goal ($x_{t_n}\in\X_G$) and avoids all obstacles ($x_t\notin\Obs$ for all $t\in[0,t_n]$) for all possible initial states $x_0\in X_0$, parameters $\theta\in\Theta$, and disturbances $w_t\in \W$. Restricting the search to piecewise-constant control trajectories is common in the literature \cite{kleinbort2018,li2016asymptotic}. 
The resulting robust planning problem is stated as follows.

\vspace{1mm}

\noindent\textit{Robust Planning Problem.}
Compute a control trajectory 
$u_t=\sum_{i=0}^{n-1} \bar{u}_i \mathbbm{1}_{\left[t_i,t_{i+1}\right)}(t)$ defined by a sequence of $n\in\mathbb{N}$ constant control inputs and durations $\{(\bar{u}_0,\tau_0),{\dots},\\(\bar{u}_{n-1},\tau_{n-1})\}$  with $t_0=0$, $\smash{t_i=\sum_{j=0}^{i-1}\tau_j}$ for $i=1,\dots,n$ 
such that the corresponding uncertain state trajectory defined from \eqref{eqn:system_dynamics}
as
\begin{align}\label{eq:reachable_state}
    {}^{\theta,w} x_t^u(x_0) \triangleq x_0 + \int_{0}^{t} {}^{\theta}f\left(x_\tau, u_\tau, w_\tau\right)\d\tau,\
\end{align}
satisfies, for all $x_0\in X_0$, all $\theta\in\Theta$, and all $w_{t\in[0,t_n)}\in\W$,
\begin{align}\label{eq:constraints}
    {}^{\theta,w}x^u_{t_n} \in \X_G,
    \ \ \text{and}\ \ \ 
    {}^{\theta,w}x^u_t \notin \Obs\ \text{ for all } 0\leq t\leq t_n.
\end{align}
Note that $\theta$ corresponds to epistemic uncertainty of the system (e.g., uncertain drag coefficients, see Section \ref{sec:experiments}). This uncertainty is propagated along the entire state trajectory, which makes it particularly challenging to address. One could generalize this formulation to allow for time-varying parameters $\theta_t$ and state-dependent disturbances $w_t(x_t)$; we leave such extensions to future work. Below, we discuss generalization to feedback-controlled systems and hybrid systems.
\\[1mm]
\noindent
\textbf{Feedback controllers} can be accounted for in the problem formulation above to reduce uncertainty and improve performance as follows. 
Given a pre-specified feedback controller  $\kappa:\X\times\U\rightarrow\U$, planning with the closed-loop dynamics $\dot{x}_t\triangleq {}^{\theta}f(x_t,\kappa(x_t,u_t),w_t)$ enables accounting for the effect of feedback at run-time. Our formulation and algorithm are general and agnostic to the choice of feedback controller $\kappa$; we refer to Section \ref{sec:experiments} for examples. In the subsequent sections, we abstract away $\kappa$ and denote the closed-loop dynamics with ${}^\theta f(x,u,w)$.
\\[1mm]
\noindent
\textbf{Hybrid Systems.}
The planning problem can be generalized to systems with hybrid dynamics
\vspace{-4mm}
\begin{subequations}
\begin{align}
    \dot{x}_t &={}^{\theta}f(x_t,u_t,w_t,\sigma_t), (x_t,u_t,w_t) \not \in {}^{\theta}\mathcal{G}, \\
    \left(\sigma_t^+,x_t^+ \right)&={}^{\theta}r(x_t,u_t,w_t,\sigma_t), (x_t,u_t,w_t) \in {}^{\theta}\mathcal{G},
\end{align}
\end{subequations}
where $\sigma \in \Sigma$ denotes the system mode, $\Sigma\subset\N$ is a finite set, ${}^{\theta}\mathcal{G}$ is the zero-measured set of guards where mode transitions occur, and  $r$ is the mode transition dynamics on the guards. We refer to \cite{wu2020r3t} for further details.

\vspace{-3mm}
\section{\robustrrt}
\label{sec:robust_rrt_algorithm}
\vspace{-2mm}

Solving the robust planning problem requires an algorithm that is capable of both (1) finding control inputs that steer the system to the goal $\X_G$, and (2) ensuring that all constraints are satisfied for all possible uncertain parameters and disturbances. To solve this problem, we propose \robustrrt, an algorithm that combines sampling-based motion planning and reachability analysis.

\textit{Forward reachability analysis} is the key technique used in this work to propagate uncertainty along the state trajectory. 
Given a control trajectory $u$ and a set of initial states $X_0$, we define the \textit{reachable set} of \eqref{eqn:system_dynamics} at time $t\geq0$ as
\begin{equation}
\label{eq:reachable_set}
X_t^u(X_0) \triangleq
\{{}^{\theta,w} x_t^u(x_0)=\eqref{eq:reachable_state}, \  \theta\in\Theta, 
w_t\in\W, x_0\in X_0\}.
\end{equation}
Using reachable sets, the robust planning problem is equivalent to %
finding a control input and duration sequence $\{(\bar{u}_0,\tau_0),\dots,(\bar{u}_{n-1},\tau_{n-1})\}$
such that 
\begin{align}\label{eq:reach_set_valid}
    X_{t_n}^u(X_0)\subseteq  \X_G,
    \ \text{and}\ 
    X_{t}^u(X_0) \cap \Obs = \emptyset \ \text{for all }  0\leq t\leq t_n.
\end{align}

We leverage \rrt(\cite{lavalle_2001}) to tackle the planning problem. %
The key data structure in \rrt is a \textit{tree} $T=(V,E)$, where $V=\smash{\{v_i\}_{i=1}^{|T|}}$ is a set of \textit{nodes}, $E=\smash{\{e_{i_j\rightarrow i'_j}\}_j}$ is a set of directional \textit{edges}, and 
each $i'$th node $v_{i'}$ is connected to its unique parent node $v_{i}$ with edge $e_{i\rightarrow i'}$.  
Each node in an \rrt corresponds to a state $v_i\triangleq\{x_i\}$, and edges between nodes represent trajectories obtained for a choice of control-duration pair $(\bar{u}_i,\tau_i)$. From the starting \textit{root node} $v_0\triangleq\{x_0\}$, an \rrt grows by randomly selecting nodes in $V$ to extend from with sampled controls and durations.  
In this work, we generalize \rrt to the robust setting. We present \robustrrt in Algorithm \ref{alg:robust_rrt}, and describe its main features below. 
\begin{figure*}[t]
\vspace{-8mm}
\begin{minipage}{0.53\linewidth}
    \begin{algorithm}[H]%
    \caption{\robustrrt}%
    \begin{algorithmic}[1]
    \Require{$i_{max}$, $X_0$, $\X_G$, $\Obs$}
    \Ensure{$\{(\bar{u}_0,\tau_0),\dots,(\bar{u}_{n-1},\tau_{n-1})\}$}
    \State{Initialize $T, \hat{T}$ with $X_0$ and  $\hat{x}_0\in X_0$}
    \For{$i=0$ to $i_{max}$}
        \label{algline:rrt_iteration}
    	\State{$X_s, \hat{x}_s \gets \texttt{sample_node}(\X,T,\hat{T})$}
        \label{algline:rrt_node_sampling:node}
    	\State{$(\bar{u}, \tau)\sim\textrm{Unif}(\U)\times\textrm{Unif}([0,\tau_{\textrm{max}}])$}
        \label{algline:rrt_node_sampling:control}
        \State{$\hat{x}_{new} \gets \hat{x}_s + \int_0^\tau {}^{\hat{\theta}}f(x_{\tau'}, \bar{u}, \hat{w}_{\tau'})\d\tau'$\hspace{-1mm}}
        \label{algline:nominal_state_compute}
        \State{$X_{new}\gets$  \texttt{compute_reach_set}($X_s, \bar{u}, \tau$)}
        \label{algline:compute_reach_set}
        \If{\scalebox{0.95}{\texttt{collision}$(\Obs,X_{new})\text{ or } X_{new}{=}\emptyset$}}
    	\label{algline:collision_check}
    	\State{\textbf{continue}}
    	\EndIf
        \label{algline:add_node}
        \State{Add $X_{new}, \hat{x}_{new}$ to $T, \hat{T}$}
        \If{$X_{new} \subseteq  \X_{G}$}
            \label{algline:goal_check}
            \State{\textbf{return} \texttt{BuildPath}($T, X_{new}$)}
        \EndIf
    \EndFor
    \State{\textbf{return} $\emptyset$}
    \end{algorithmic}
    \label{alg:robust_rrt}
    \end{algorithm}
\end{minipage}
\hspace{0.01\linewidth}
\begin{minipage}{0.45\linewidth}
    \begin{algorithm}[H]
    \caption{\texttt{sample_node()} }
    \begin{algorithmic}[1]
    \Require{$\X$, $T$, $\hat{T}$, $\zeta$}
    \Ensure{$X_s, \hat{x}_s$}
    \State{Sample $x_s\sim \textrm{Unif}(\X)$}
    \label{algline:state_sampling}
    \State{$d\gets{\min_{x \in \hat{T}}}\|x-x_s\|$}
    \If{$d>\zeta$}
    \State{$\hat{x}_s\gets{\argmin_{\hat{x} \in \hat{T}}}\|\hat{x}-x_s\|$}
    \Else
    \State{$\hat{x}_s\,{\sim}\, \textrm{Unif}(\{x\,{\in}\, \hat{T} \,{:}\, \|x\,{-}\,x_s\|\,{\leq}\, \zeta\})$}
    \label{algline:node_tie_breaking}
    \EndIf
    \State{$X_s\gets X\in T$ corresponding to $\hat{x}_s$}
    \State{\textbf{return} $(X_s,\hat{x}_s)$}
    \end{algorithmic}
    \label{alg:sample_node}
    \end{algorithm}
\end{minipage}
\vspace{-8mm}
\end{figure*}
\\[-3mm]
\noindent
\textbf{Reachable sets as \rrt nodes:}
Instead of individual states, each node $v_i$ of the tree $T=(V,E)$ in \robustrrt corresponds to a reachable set $v_i\triangleq\left\{X_{t_i}^u(X_0)\right\}$. The initial node $v_0\triangleq \left\{X_0\right\}$ corresponds to all possible initial states of the system. 
Each edge $e_{i\rightarrow i'}$ is defined by a control input-switching time pair $(\bar{u}_i,\tau_i)$ and connects two reachable-set nodes according to \eqref{eq:reachable_set}. 
A valid node $v_i\in T$ should be collision-free, i.e., satisfy $v_i \cap \Obs = \emptyset$ in \eqref{eq:reach_set_valid}; we include this check in \robustrrt within the \texttt{collision()} subroutine in Algorithm \ref{alg:robust_rrt}. We note that the forward simulation and collision checking (lines \ref{algline:compute_reach_set} and \ref{algline:collision_check}) timestep size may be smaller than $\tau_i$ for improved accuracy. If a reachable-set node $v_n=X_{t_n}^u(X_0)$ is fully contained in the goal region $\X_G$, \robustrrt terminates and returns a control input and duration sequence that solves the robust planning problem.
\\[1mm]
\noindent
\textbf{Node selection with a nominal tree:}
In addition to the reachable set tree $T$, \robustrrt also constructs a \textit{nominal tree} $\hat{T}=(\hat{V}, \hat{E})$ akin to standard \rrt. Each node $\hat{v}_i\in\hat{T}$ corresponds to a nominal state $\hat{v}_i = \{\hat{x}_{t_i}\}$, $\smash{\hat{x}_{t_i} \triangleq \hat{x}_0} + \smash{\int_0^{t_i} {}^{\hat{\theta}}f(\hat{x}_{\tau'}, u_{\tau'}, \hat{w}_{\tau'})\d {\tau'}}
$, where the nominal parameters $\hat\theta\in\Theta$ and disturbances $\hat{w}_s\in\W$ are selected prior to running \robustrrt. %
The nominal tree $\hat{T}$ is grown in parallel to the robust tree $T$ so that each nominal node $\hat{v}_i\in\hat{T}$ uniquely corresponds to a node $v_i\in T$. 
We describe our node selection strategy in Algorithm \ref{alg:sample_node} and Fig. \ref{fig:node_sampling}. 
Given a hyperparameter $\zeta>0$ (that can be chosen arbitrarily small), a state $x_s$ is first sampled uniformly from the state space $\X$. Next, we select the nominal state $\hat{x}_s\in\hat{T}$ that is the closest to the sample $x_s$ if all distances $\|\hat{x}_s-x_s\|$ are larger than $\zeta$. Otherwise, we randomly choose any $\zeta$-close nominal state $\hat{x}_s\in\hat{T}$. 
Then, to extend the robust tree $T$, we select the reachable-set node $X_s\in T$ that uniquely corresponds to $\hat{x}_s\in\hat{T}$. 
In contrast to \cite{panchea2017extended,pepy2009reliable}, this approach circumvents the need to compute distances to reachable sets for nearest-neighbor search and still preserves Voronoi bias to some degree for rapid exploration.
\\[1mm]
\noindent
\textbf{Control and duration selection:}
For simplicity in proving probabilistic completeness (PC) but without loss of generality, we sample the control $\bar{u}$ and duration $\tau$ from independent uniform sampling distributions over $\mathcal{U}$ and $[0, \tau_{max})$. %We note that in order to stabilize the system, $\bar{u}$ can be a parameterization of some predetermined closed loop control law $\kappa(x, \bar{u})$.
\vspace{-3mm}
\section{\robustrrt is Probabilistically Complete (PC)}
\label{sec:pc_proof}
\vspace{-2mm}
Before providing our main theoretical result, we state our assumptions about the dynamics of the system. We recall that in the presence of a stabilizing controller, ${}^\theta f(x,u,w)$ denotes the closed-loop dynamics of the system, see Section \ref{sec:problem_formulation}. % For ease of notation, we absorb any (predetermined) closed-loop controller into the system dynamics, i.e. ${}^{\theta}f(x,u,w)$ is the closed loop dynamics.
\begin{assumption}[Lipschitz continuous bounded dynamics]
\label{assumption:lipschitz_bounded}
\vspace{-2mm}
The dynamics in \eqref{eqn:system_dynamics}  are Lipschitz continuous and bounded, i.e., there exist two constants $K,\lambda \geq 0$ such that for any $w\in\W$, any $\theta\in\Theta$, 
any $x,y\in\R^n$, and any $u,v\in\U$,%
\begin{equation}
\label{eqn:lipschitz_continuity}
    \|{}^{\theta}f(x,u,w)-{}^{\theta}f(y,v,w)\|\leq K (\|x-y\|+\|u-v\|).
\end{equation}
\vspace{-4mm}
\begin{equation}
\label{eq:bounded}
\sup_{(x,\theta,u,w)\in\mathbb{R}^n\times\Theta\times\mathcal{U}\times\mathcal{W}}\|{}^{\theta}f(x,u,w)\| \leq \lambda,
\end{equation}
\end{assumption}
\noindent Assumption \ref{assumption:lipschitz_bounded} is mild, holds for a wide class of dynamical systems found in robotics applications, and is required by previous RRT PC proofs, see \cite{kleinbort2018,li2016asymptotic}. 
Lipschitz continuity for all uncertain parameters and disturbances in \eqref{eqn:lipschitz_continuity} is a typical smoothness assumption that ensures the existence and uniqueness of solutions to the ODE in \eqref{eqn:system_dynamics}. The boundedness of the dynamics in \eqref{eq:bounded} allows bounding the evolution of the state trajectory over time. Since $f$ is assumed to be continuous, \eqref{eq:bounded} is automatically satisfied if the system evolves in a bounded operating region (i.e., if $\X$ is compact), which is a common situation in robotics applications. We leave the PC proof for hybrid systems to future work. 

Next, we make an assumption about the structure of the planning problem. As computing a safe plan under bounded uncertainty amounts to ensuring sufficient separation between sets, we use the Hausdorff distance metric $d_H$ to describe distances between any two nonempty compact sets $X_1, X_2\subset\X$ as
\begin{align}\label{eq:hausdorff_distance}
d_H(X_1, X_2) \triangleq 
\max\Big\{
&\sup_{x_1 \in X_1}
\Inf_{x_2 \in X_2}
\|x_1 - x_2\|,
\sup_{x_2 \in X_2}
\Inf_{x_1 \in X_1}
\|x_1 - x_2\|
\Big\}.\nonumber
\end{align}
\label{def:hausdorff}
With this metric, we extend the typical $\delta$-clearance assumption from the sampling-based motion planning literature \cite{karaman2011sampling,kleinbort2018,li2016asymptotic} to the robust problem formulation.
\begin{assumption}[Existence of a $\delta$-robust solution]
\label{assumption:existence_robustly_feasible}
\vspace{-2mm}
There exists a constant $\delta>0$ and an $n$-step piecewise-constant control trajectory 
$u^*_t=\sum_{i=0}^{n-1} \bar{u}^*_i \mathbbm{1}_{\left[t^*_i,t^*_{i+1}\right)}(t)$ defined by a control input and duration sequence $\{(\bar{u}^*_0,\tau^*_0),\dots,(\bar{u}^*_{n-1},\tau^*_{n-1})\}$ with $t_0^*=0$, $t^*_i=\sum_{j=0}^{i-1}\tau^*_j$ for $i=1,\dots,n$ 
such that the associated reachable set trajectory $X_t^{u^*}(X_0)$ with $t\in[0,t_n^*]$ solves the robust planning problem:
\begin{align}
    X_{t^*_n}^{u^*}(X_0)\subseteq  \X_G,
    \ \text{and}\ 
    X_{t}^{u^*}(X_0) \cap \Obs = \emptyset \ \text{for all }  0\leq t\leq t^*_n.
\end{align}
Moreover, any control trajectory $u$ with associated reachable sets $X_t^{u}(X_0)$ such that
\vspace{-4mm}
\begin{align}\label{eq:control_with_reachsets_delta_close_to_solution}
    d_H(X^u_{\sigma^u(t)}(X_0), X^{u^*}_t(X_0)) \leq \delta \ \text{for all }  0\leq t\leq t^*_n
\end{align}
for some time reparameterization $\sigma^u:[0,t_n^*]\rightarrow [0,t_n]$ 
also solves the problem:
\begin{align}\label{eq:feasible_solution}
    X_{\sigma^u(t_n)}^u(X_0)\subseteq  \X_G,
    \ \text{and}\ 
    X_{\sigma^u(t)}^u(X_0) \cap \Obs = \emptyset \ \text{for all }  0\leq t\leq t^*_n.
\end{align}
\end{assumption}
\noindent This assumption states that there exists a solution $u^*$ whose associated reachable sets $X^{u^*}_{t}$ is at least $\delta$ (Hausdorff) distance away from all obstacles $\Obs$ and $\delta$ inside $\X_G$.
We highlight the explicit dependency of the reachable sets on $u$ to facilitate our subsequent proofs. Indeed, by Assumption \ref{assumption:lipschitz_bounded}, small differences in control inputs translate into small Hausdorff distances between reachable sets.
\begin{lemma}[Reachable sets are Lipschitz]
\label{lemma:lipschitz_continuity_all}
Let $X_0^1,X_0^2\subset\mathbb{R}^n$ be two nonempty compact sets and 
$u^1,u^2$ be two control inputs of durations $t_1,t_2\geq 0$, respectively, such that $u_t^i=\bar{u}_i$ for all $t\in[0,t_i)$ with $i=1,2$.  
Then, under Assumption \ref{assumption:lipschitz_bounded}, there exists a bounded constant $L\geq 0$ such that
\begin{equation}
\label{eq:reachable_set_deviation_full}
d_{\textrm{H}}(X_{t_1}^{u^1}(X_0^1),X_{t_2}^{u^2}(X_0^2))
\leq
L(d_{\textrm{H}}(X_0^1,X_0^2)+|t_1-t_2|+\|\bar{u}_1-\bar{u}_2\|).
\end{equation}
\end{lemma}
\noindent We refer to the Appendix \cite{Wu2022robustrrt} for a proof. 
With Lemma \ref{lemma:lipschitz_continuity_all} and Assumption \ref{assumption:existence_robustly_feasible}, a sufficient condition to proving PC consists of showing that \robustrrt eventually finds an $n$-step piecewise control trajectory $u$ that is close enough to $u^*$\footnote{Note that the control length $n$ of the control $u^*$ in Assumption \ref{assumption:existence_robustly_feasible} is unknown to \robustrrt. In practice, the algorithm may discover other valid solutions.}. Notably, our results do not require Chow's condition \cite{chow1940systeme,li2016asymptotic} and use an argument that is similar to \cite{Papadopoulos2014} but applies to the robust planning problem setting. Next, we prove that \robustrrt is PC. 
\begin{lemma}[\robustrrt eventually extends from any node with the right input and duration]%
\label{lemma:almost_surely_extension}
Let $T_m$ denote the reachable set tree of \robustrrt at iteration $m$ of Algorithm \ref{alg:robust_rrt}, and $v_i\in T_m$ be any reachable-set node. 
Let  $\epsilon>0$, $\bar{u}^*\in\U$, and $\tau^*\in[0,\tau_{\textrm{max}}]$ be arbitrary. 
Then, almost surely, \robustrrt eventually extends from the node $v_i$ with an input $\bar{u}\in\U$ and a duration $\tau\in[0,\tau_{\textrm{max}}]$ such that
$
|\tau-\tau^*|+\|\bar{u}-\bar{u}^*\| \leq \epsilon.
$
\end{lemma}

\begin{proof}
Let $|T_m|=|\hat{T}_m|$ denote the size of the \robustrrt trees at iteration $m\in\N$. Denote by $N_m$ the event that \robustrrt selects the node $v_i$ to extend from, 
by $C_m$ the event that $\robustrrt$ samples the control $\bar{u}$ and duration $\tau$ such that $
|\tau-\tau^*|+\|\bar{u}-\bar{u}^*\| \leq \epsilon$, 
and by $E_m$ the event where both $C_m$ and $N_m$ happen simultaneously, i.e.,
$E_m \triangleq C_m\cap N_m.$
We denote the event such that $\robustrrt$ \textit{eventually} extends from $v_i$ with the desired control-duration pair $(\bar{u},\tau)$ as
$
E \triangleq \bigcup_{k=0}^\infty E_{m+k}.
$
To prove Lemma \ref{lemma:almost_surely_extension}, we show that
$\Prob\left(E\right) = 1$. 

By our control and duration selection strategy,  
$
\Prob(C_m)=c>0
$
for some strictly positive constant %
$c$ that is independent of $m$,  which follows from using a uniform distribution over $\U\times[0,\tau_{\textrm{max}}]$ at each iteration $m$ of the algorithm. %
Using the definition of \texttt{sample_node()}, we show that
$
\Prob(N_m) \geq b/|T_m|
$
for some strictly positive constant $b \triangleq \mathbb{P}(x_s \in \mathcal{B}_\zeta(\hat{x}_i))>0$, where $\hat{v}_i=\{\hat{x}_i\}\in\hat{T}_m$ is the nominal node that corresponds to the reachable-set node $v_i\in T$. 
Observe that under \texttt{sample_node()}, a sufficient condition for choosing to extend from $v_i$ is (1) to sample a state $x_s$ that is $\zeta$-close to $\hat{x}_i$, and (2) to select $\hat{x}_i$ (after uniform tie-breaking) among all states in $\mathcal{B}_\zeta(x_s)$. The probability of (1) is $\mathbb{P}(x_s \in \mathcal{B}_\zeta(\hat{x}_i))$, which is strictly positive since the support of the sampling scheme covers $\mathcal{X}$ (Line \ref{algline:state_sampling} of Algorithm \ref{alg:sample_node}). The probability of (2) is lower bounded by the worst case scenario where all the nominal nodes $\hat{v}\in\hat{T}_m$ belong to $\mathcal{B}_\zeta(x_s)$. In the worst case, \texttt{sample_node()} simply perform uniform selection among the $|\hat{T}_m| = |T_m|$ nominal nodes (Line \ref{algline:node_tie_breaking} of Algorithm \ref{alg:sample_node}), which is at least $1/|T_m|$.

Third, $N_m$ and $C_m$ are independent events, by definition of  \texttt{sample_node()} and since the control-duration pair $(\bar{u},\tau)$ is sampled independently. %
Thus, 
$
    \Prob(E_m)
    \\\triangleq
    \Prob(C_m\cap N_m) =  \Prob(C_m)\Prob(N_m) \geq \frac{c b}{|T_m|}.
$ 
At each iteration, at most one node is added to the tree of \robustrrt, so $|T_{m+1}|\leq |T_m|+1$. Hence,
\begin{equation}
\label{eqn:prob_successful_extension_indep}
\Prob(E_{m+k})
\geq \frac{c b}{|T_{m+k}|} \geq \frac{c b}{|T_m|+k}.
\end{equation}

Finally, proving Lemma \ref{lemma:almost_surely_extension} amounts to proving $\Prob(E)=1$. By De Morgan's law, this is equivalent to proving 
$
\Prob\left(\bigcap_{k=0}^\infty E_{m+k}^c\right)=0.
$
From \eqref{eqn:prob_successful_extension_indep},
$
    \Prob(E^c_{m+k}) = 1-\Prob(E_{m+k}) \leq 1-\frac{c b}{|T_m|+k}.
$ 
Therefore,
\begin{align}
\label{eqn:robustrrt_failure_probability}
\Prob\left(\bigcap_{k=0}^{\infty} E_{m+k}^c\right) 
&\leq   
\prod_{k=0}^{\infty}
 \Prob(E^c_{m+k})
\leq
\prod_{k=0}^{\infty}\left(1-\frac{c b}{|T_m|+k}\right)
    =
    0.
\end{align}
The last equality follows from Equation 9 of \cite{wu2020r3t}. This concludes the proof.%
\end{proof}

\begin{theorem}[\robustrrt is PC]
\label{thm:randup_rrt_pc} 
Under Assumptions \ref{assumption:lipschitz_bounded} and  
\ref{assumption:existence_robustly_feasible}, 
\robustrrt almost surely solves the robust planning problem, i.e., 
it almost surely finds an $n$-step control and duration sequence $\left\{(\bar{u}_0, \tau_0),\dots,(\bar{u}_{n-1}, \tau_{n-1})\right\}$ such that the resulting reachable sets $\left\{X_t^u(X_0), t\in[0,t_n]\right\}$ satisfy \eqref{eq:reach_set_valid}.
\end{theorem}
\begin{figure}[!t]
\vspace{-5mm}
    \centering
  \begin{minipage}[c]{0.45\textwidth}
  \vspace{-1mm}
    \includegraphics[width=1\textwidth]{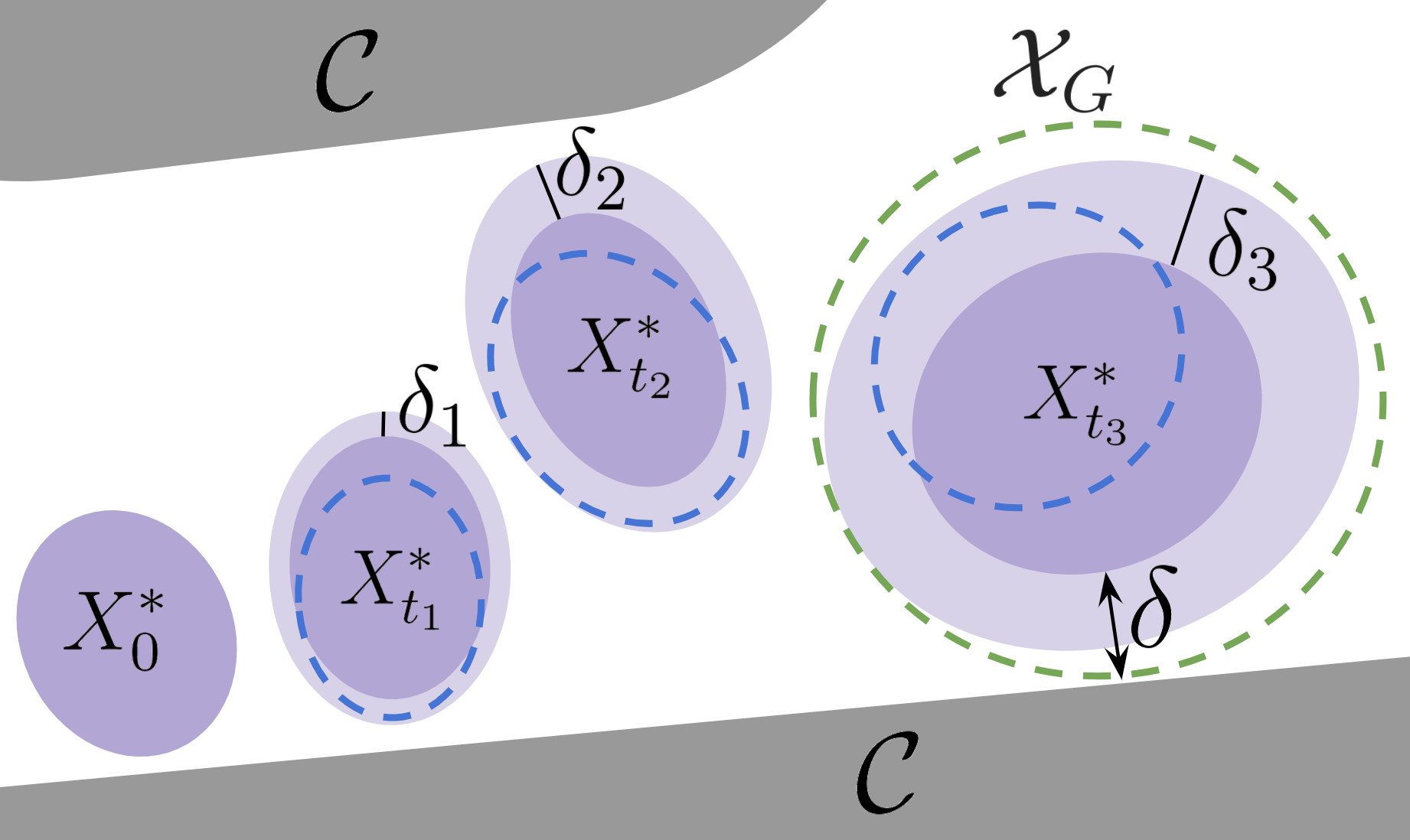}
    \end{minipage}\hfill
  \begin{minipage}[c]{0.45\textwidth}
      \captionof{figure}{Illustration of the growing funnel argument with $\delta_{i+1}=(2\max(L, 1))\delta_i$ for the proof of Theorem \ref{thm:randup_rrt_pc}. Note that $\delta_i \leq \delta$, and $\delta_i$ increases with $i$. The blue dashed regions show a possible path returned by \robustrrt. The light purple region for each $X_{t_i}^*$ illustrates the set $\left\{x|d_H(\{x\}, X_{t_i}^*) \leq \delta_i \right\}$.}
    \label{fig:growing_funnel_argument}
    \end{minipage}
\vspace{-10mm}
\end{figure}
\begin{proof}
\vspace{-5mm}
We proceed in two steps. First, we show that \robustrrt almost surely finds an $n$-step control trajectory $u$ that is close enough to the robust solution $u^*$ from Assumption \ref{assumption:existence_robustly_feasible}. 
We then show that such a trajectory solves the problem. 

\textbf{Step 1:} Let\footnote{Choosing $\delta_i=\frac{\delta i}{n}$ would not suffice for the proof. Indeed, the error accumulated from previous time steps grows exponentially at a rate bounded by the Lipschitz constant. Nevertheless, since the trajectory has finitely many steps, the error is still bounded.} $\delta_i \triangleq \frac{\delta}{(2\max(L, 1))^{n-i}}$
for any $i=0,\dots,n$, so  $\delta_{i+1}=(2\max(L, 1))\delta_i\\ \geq 2L\delta_i$ and  $\delta_{i+1}\leq\delta$ for all $i=0,\dots,n-1$.  
We show that
\robustrrt almost surely discovers an $n$-step control trajectory $u$ with reachable sets $X_{t_i}$ such that
\begin{equation*}
    |\tau_i-\tau_i^*|+\|\bar{u}_i-\bar{u}_i^*\| \leq \delta_i
    \ \ \text{ and } \ \ 
    d_H(X_{t_i}, X_{t^*_i}^*) \leq \delta_i 
    \ \ \text{for all} \ \ 
    i=0,\dots,n.
\end{equation*}
To prove this claim, we note that the tree $T$ initially only contains the starting reachable set $X_0$, so that $X_{t_0}=X_{t^*_0}^*=X_0$. Therefore, $d_H(X_{t_0}, X_{t^*_0}^*) = 0 \leq \delta_0$.
By induction, we assume that at step $i$, the tree $T$ of \robustrrt contains a reachable set $X_i$ that satisfies $d_H(X_{t_i}, X_{t_i^*}^*) \leq \delta_i$.
Then, we wish to prove that \robustrrt almost surely extends from $X_{t_i}$ to $X_{t_{i+1}}$ with some $(\bar{u}_i,\tau_i)$ such that  
$
|\tau_i-\tau_i^*|+\|\bar{u}_i-\bar{u}_i^*\| \leq \delta_i
$ and such that $\smash{d_H(X_{t_{i+1}}, X_{t_{i+1}^*}^*)} \leq \delta_{i+1}$. 
By choosing $\epsilon\triangleq \delta_i$ in Lemma \ref{lemma:almost_surely_extension}, we know that given $\delta_i$, almost surely, \robustrrt will select $v_i$ to extend from and some $(\bar{u}_i,\tau_i)$ such that  
$
|\tau_i-\tau_i^*|+\|\bar{u}_i-\bar{u}_i^*\| \leq \delta_i.
$
Using \eqref{eq:reachable_set_deviation_full}, the resulting reachable set $X_{t_{i+1}}$ satisfies
\begin{align*}
d_H(X_{t_{i+1}}, X_{t^*_{i+1}}^*) 
&\leq
L(d_{\textrm{H}}(X_{t_{i}}, X_{t^*_{i}}^*)+|\tau_i-\tau_i^*|+\|\bar{u}_i-\bar{u}_i^*\|),
\leq
L(\delta_i+\delta_i) \leq
\delta_{i+1}.
\end{align*}
Iterating over $i=0,\dots,n-1$, this concludes the proof of Step 1.

\textbf{Step 2.} Consider an $n$-step control trajectory $u$ such that $
|\tau_i-\tau_i^*|+\|\bar{u}_i-\bar{u}_i^*\| \leq \delta_i
$ for all $i=0,\dots,n-1$, and $d_H(X_{t_i}, X_{t^*_i}^*) \leq \delta_i$ for all $i=0,\dots,n$. The goal is to show that this trajectory is feasible for the robust planning problem. %
From \eqref{eq:reachable_set_deviation_full}, for all times  $t\in [0,\tau_i)$, we have
\begin{align*}
d_H(X_{t_i+t}^u(X_0), X_{t_i^*+t}^{u^*}(X_0)) 
&\leq
L(d_{\textrm{H}}(X_{t_i},X_{t_i^*}^*)+\|\bar{u}_i-\bar{u}_i^*\|)
\leq \delta_{i+1}\leq \delta.
\end{align*}
Thus, $X_t^u(X_0)$ satisfy  \eqref{eq:control_with_reachsets_delta_close_to_solution} in Assumption \ref{assumption:existence_robustly_feasible}, and $u$ indeed solves the problem.

Combining Steps 1-2 concludes the proof of Theorem \ref{thm:randup_rrt_pc}.
\end{proof}

\paragraph{Extensions to other set-based RRTs.} 
Our PC proof relies on bounding the Hausdorff distance between sets via a valid ball of control and duration samples. This proof technique can be extended to other set-based \rrt methods by replacing \texttt{compute_reach_set()} with a set-valued map $F(X_0, u, t)$. For instance, in \robustrrt, $F(X_0, u, t) \triangleq X_t^u(X_0)$; padding a nominal trajectory $\hat{x}^u$ with an invariant-set $X_{\textrm{inv}}$ corresponds to $F(X_0, u, t)\triangleq\hat{x}_t^u\oplus X_{\textrm{inv}}$. To generalize Theorem \ref{thm:randup_rrt_pc}, one should replace \eqref{eq:reachable_set_deviation_full} in Lemma \ref{lemma:lipschitz_continuity_all} with 
\begin{align*}
d_{\textrm{H}}(F(X_0^1, u^1, t_1),F(X_0^2, u^2, t_2))
\leq
L(d_{\textrm{H}}(X_0^1,X_0^2)+|t_1-t_2|+\|\bar{u}_1-\bar{u}_2\|),
\end{align*}
and replace \eqref{eq:feasible_solution} in Assumption \ref{assumption:existence_robustly_feasible} with the more conservative assumption 
$$
    F(X_0, u, \sigma^u(t_n^*))\subseteq  \X_G,
    \ \text{and}\ 
    F(X_0, u, \sigma^u(t)) \cap \Obs = \emptyset \ \text{for all }  0\leq t\leq t^*_n.
$$
Candidate algorithms in literature for this extension include
KDF-RRT \cite{verginis2021kdf}, polytopic trees \cite{sadraddini2019sampling}, FaSTrack+RRT \cite{herbert2017fastrack}, and LQR-Trees \cite{tedrake2010lqr}. We leave the details of this proof to future work.

\vspace{-3mm}
\section{Insights and Practical Considerations}
\label{sec:practical_implementation}
\vspace{-2mm}
To the best of our knowledge, \robustrrt is the most general PC robust sampling-based motion planner in the literature. The analysis applies to any smooth dynamical system affected by bounded aleatoric and epistemic uncertainty. %
The analysis reveals the following insights:
\\[1mm]
\noindent
\textbf{The growing funnel argument allows for minimal assumptions about the system}, namely bounded and Lipschitz continuous dynamics. Arguing the discovered solution remains within a fixed-radius funnel around the existing solution is unnecessary, as this approach typically relies on Chow's condition \cite{li2016asymptotic}. 
Moreover, through explicitly accounting for path and time dependency, the PC guarantee of \robustrrt does not require the system of interest to be stabilizable (or more generally, the existence of an invariant set). As long as a finite-time solution exists under Assumption \ref{assumption:existence_robustly_feasible}, \robustrrt will eventually find a feasible solution. This contrasts with invariant-set based robust planners in the literature (see Section \ref{sec:related_work}) that use a global error bound to pad constraints. 
% existing padding-based robust planners, which essentially assume all path realizations are withing the padded radius of each other for infinite horizon.
\\[1mm]
\noindent
\textbf{Computing the Hausdorff distance between reachable sets for tree expansion is not necessary}. Maintaining a nominal tree used for nearest neighbors search and tree expansion is sufficient for PC. Our approach still preserves Voronoi bias to some degree for rapid exploration.%
\noindent
\textbf{\randuprrt.} 
\robustrrt relies on forward reachability analysis to implement the \texttt{compute_reach_set()} subroutine. 
While exact methods for different problem formulations have been proposed in the literature, 
they typically rely on a specific parameterization of the dynamical system in \eqref{eqn:system_dynamics} (e.g., linear dynamics with additive disturbances, see \cite{Althoff2021} for a recent review). In general, computing the reachable sets in \eqref{eq:reachable_set} remains computationally expensive and requires approximations for tractability.

As a practical implementation, we propose \randuprrt, an algorithm that leverages a particle-based reachable set approximation algorithm called \textit{randomized uncertainty propagation} (\randup). \randup consists of sampling uncertainty values, evaluating their associated reachable states in \eqref{eq:reachable_state}, and taking the convex hull of the samples to approximate the convex hull of the true reachable set at each timestep \cite{lew2020samplingbased,LewJansonEtAl2022}. %We emphasize that each convex hull is calculated from reachable states of a single node. 
One may additionally pad the convex hull by a constant $\epsilon>0$ to provide finite-sample conservatism guarantees for the approximation; we refer to this algorithm as $\epsilon$-\randup  \cite{LewJansonEtAl2022}. %In practice, $\epsilon$ can be determined by checking constraint violation magnitudes of forward simulations and choosing $\epsilon$ to be larger than the violations. 
To apply \randuprrt to hybrid systems, %one may require as a design choice
we enforce that all states resulting from a control action belong to the same dynamic mode under all uncertainty realizations.  \randup is simple to implement, efficient, and applicable to a wide range of dynamical systems. 
While we leave the theoretical guarantees of \randuprrt to future work, we expect PC to hold under a stricter $\delta$-robust feasibility assumption for the convex hull of the true reachable sets (see Section \ref{sec:pc_proof}), and that \randuprrt finds a feasible solution with high probability assuming a sufficient number of samples for \randup, see \cite[Theorem 2]{LewJansonEtAl2022}.

\vspace{-3mm}
\section{Experiments}
\label{sec:experiments}
\vspace{-2mm}
We consider 3 challenging robust planning problems. %
The first is motion planning for a nonlinear quadrotor with epistemic (parametric) uncertainty. The second corresponds to box-pushing in a black-box physics simulator with aleatoric disturbances. The third system is a jumping robot hybrid system with unknown mass and guard surface locations. %
In all experiments, the robust satisfaction of collision avoidance and goal reaching constraints are verified using Monte-Carlo rollouts of the system. %
We observed that $\zeta$ tie breaking (Line \ref{algline:node_tie_breaking} of Algorithm \ref{alg:sample_node}) seldom occurs in practice, thus we set $\zeta=0$ for computational speed.
We show animated results in the accompanying video. 
Further implementation details are available in the Appendix \cite{Wu2022robustrrt} and at  \scalebox{0.8}{\url{https://github.com/StanfordASL/randUP_RRT.git}}. 
We report computation time relative to a naive Python \rrt implementation to factor out implementation-specific optimizations. %
\\[1mm]
\textbf{Nonlinear Quadrotor.}
We consider a quadrotor navigating in a cluttered planar environment. The system has a 4-dimensional state space and a fixed but unknown drag coefficient $\alpha\in[0.35,0.65]$. This nonlinear parametric uncertainty makes the problem challenging for typical robust planners as it introduces time correlations along the state trajectory. By combining \rrt with \randup, we account for this uncertainty by fixing the sampled value of $\alpha$ for each of the $100$ \randup trajectory particles. %
We also pad all obstacles and shrink the goal region by a constant $\epsilon{=}0.3$; this corresponds to planning with $\epsilon$-\randup 
\cite{LewJansonEtAl2022}. 

\begin{figure}[t!]
    \centering
        \begin{subfigure}[t]{0.36\textwidth}
        \includegraphics[width=\textwidth]{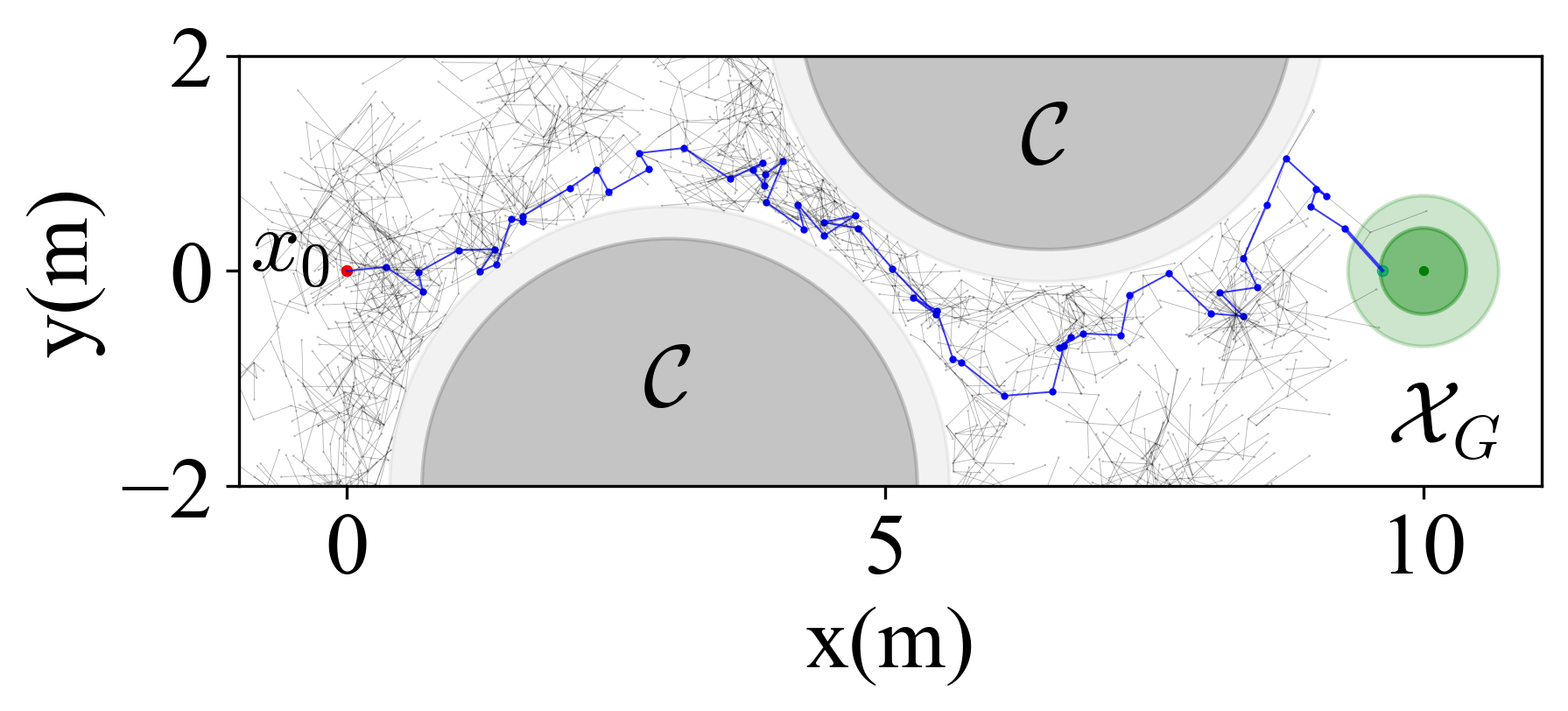}
        \caption{\rrt with $0.3$-padding.}
        \label{subfig:randup100_2obs_normal}
        \end{subfigure}
        \begin{subfigure}[t]{0.31\textwidth}
        \includegraphics[width=\textwidth,trim=59 0 0 0, clip]{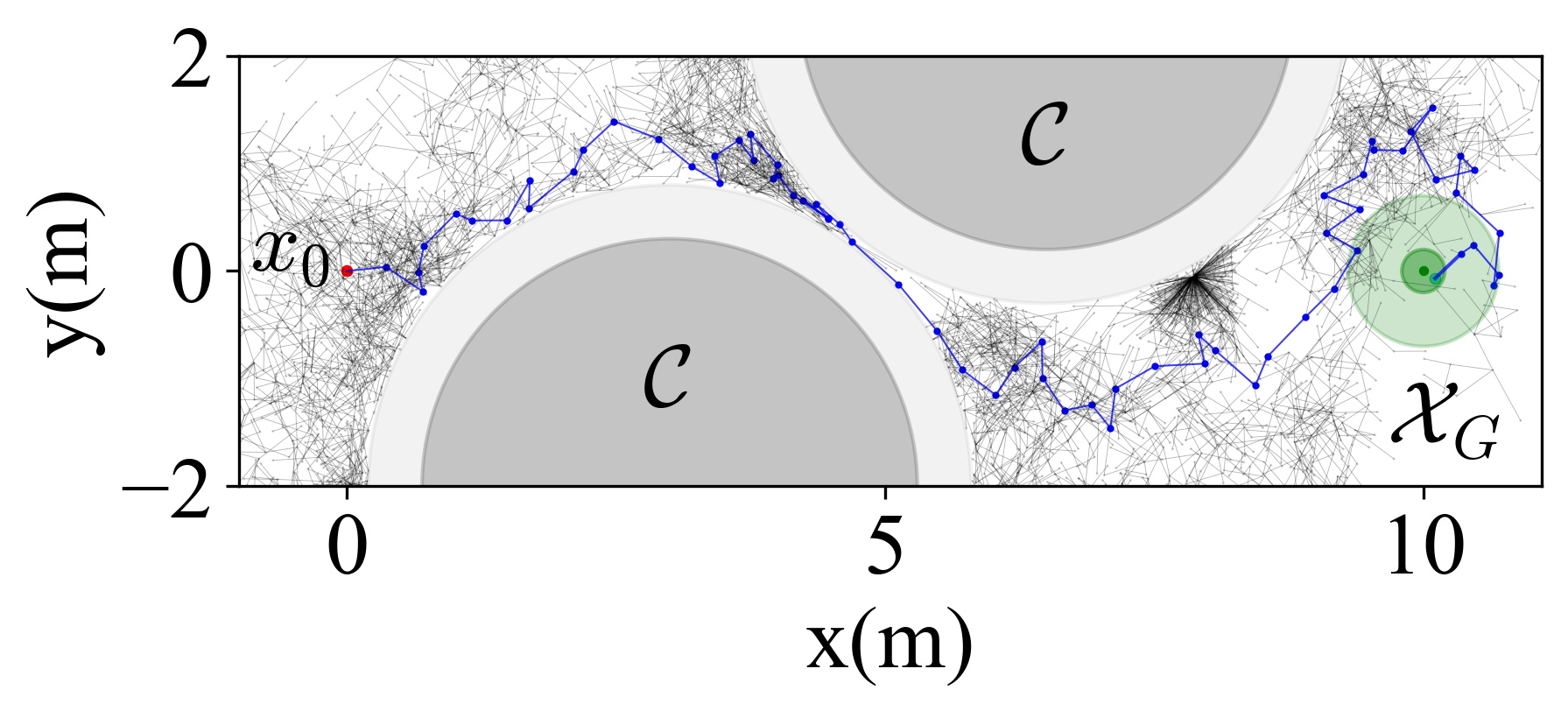}
        \caption{\rrt with $0.5$-padding.}
        \label{subfig:randup100_2obs_noisy}
        \end{subfigure}
        \begin{subfigure}[t]{0.31\textwidth}
        \includegraphics[width=\textwidth,trim=59 0 0 0, clip]{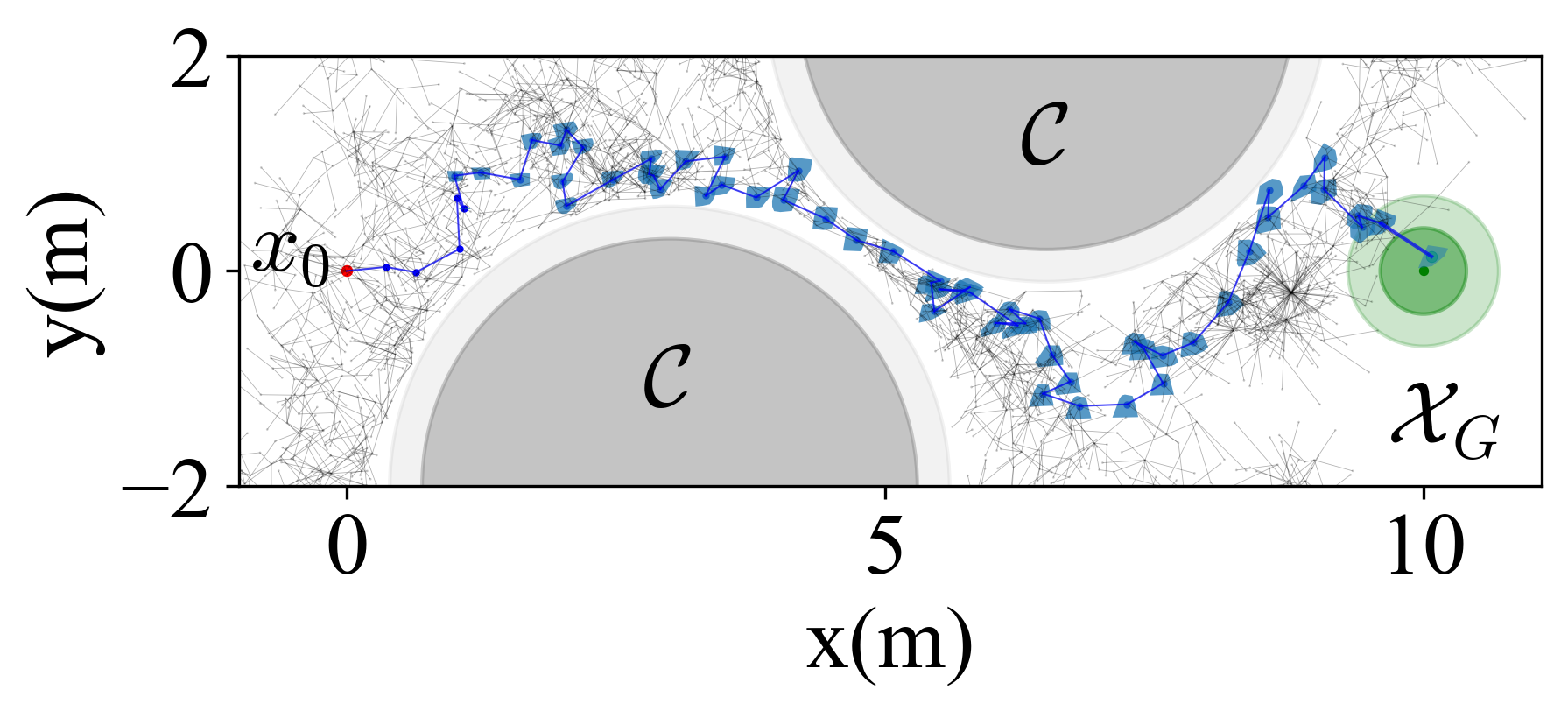}
        \caption{$\epsilon$-\randuprrt.}
        \label{subfig:randup100_3obs_noisy}
        \end{subfigure}
    \caption{Nonlinear quadrotor motion planning results. Padding is shown in light shading. Reachable sets are shown in light blue. %
    Increasing the padding constant for vanilla \rrt (\textbf{a}-\textbf{b}) increases conservatism: the problem may become artificially infeasible for large padding values. Instead, \robustrrt (\textbf{c}) directly accounts for the uncertainty of the system using reachability analysis.
    }
    \label{fig:quadrotor_plan}
    \vspace{-8mm}
\end{figure}

We summarize the results in Fig. \ref{fig:quadrotor_plan} and Table \ref{tab:quadrotor_statistics}. 
Table \ref{tab:quadrotor_statistics} indicates that simply padding constraints and relying on a standard planner to compute an approximately robust plan is insufficient to capture the worst-case ramifications of the uncertainty. By explicitly planning with reachable sets, \randuprrt is able to improve plan robustness by capturing path-dependent effects of the uncertainty. \randuprrt does indeed incur additional computational cost ($\approx 5.18\times$ of \rrt runtime), but this may be mitigated by parallelizing \randup. 

\begin{figure}[t]
    \vspace{-3mm}
    \centering
        \begin{subfigure}[t]{0.22\textheight}
        \includegraphics[width=\textwidth,trim=0 40 0 0, clip]{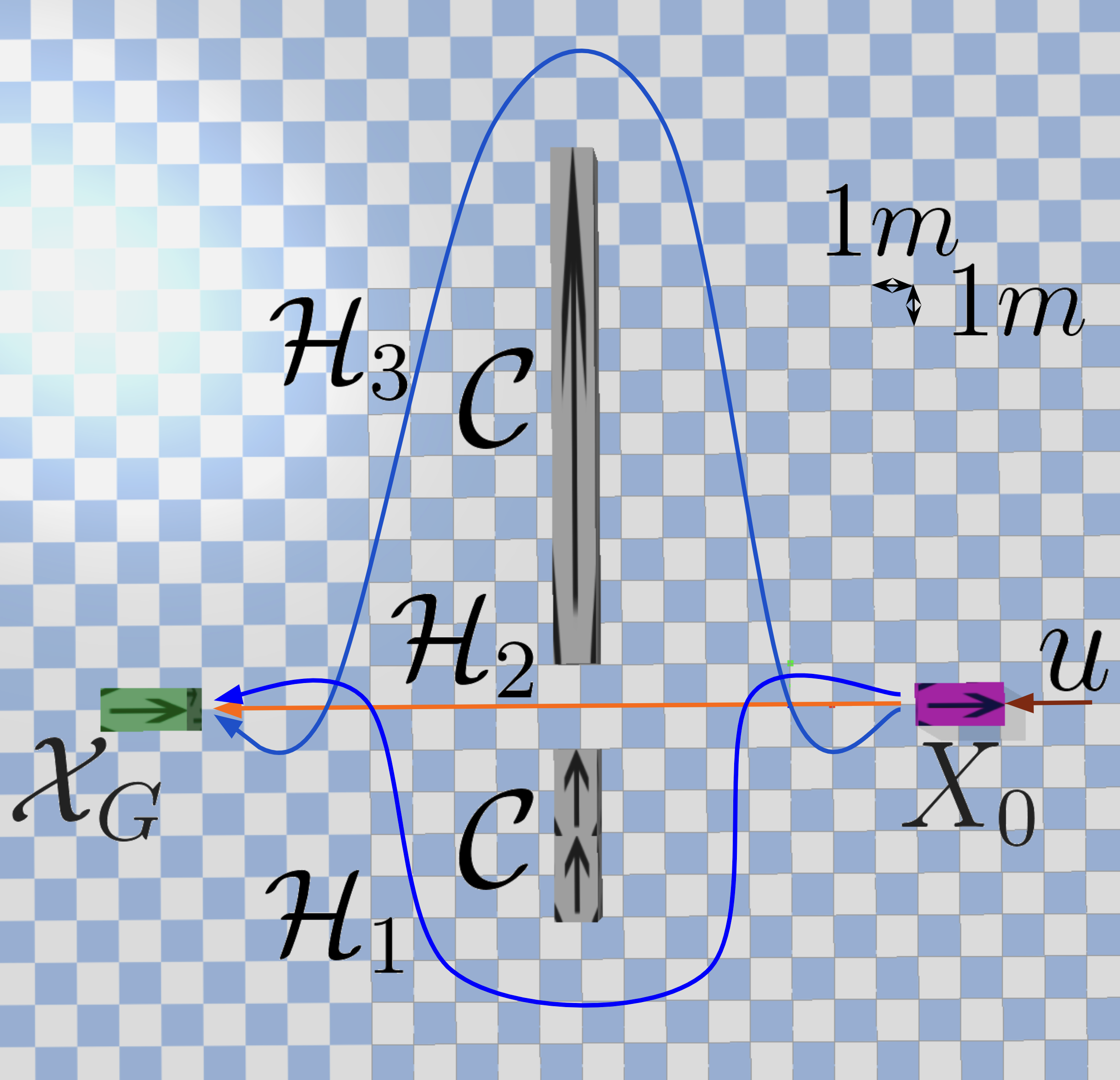}
        \end{subfigure}
        \begin{subfigure}[t]{0.22\textheight}
        \includegraphics[width=\textwidth]{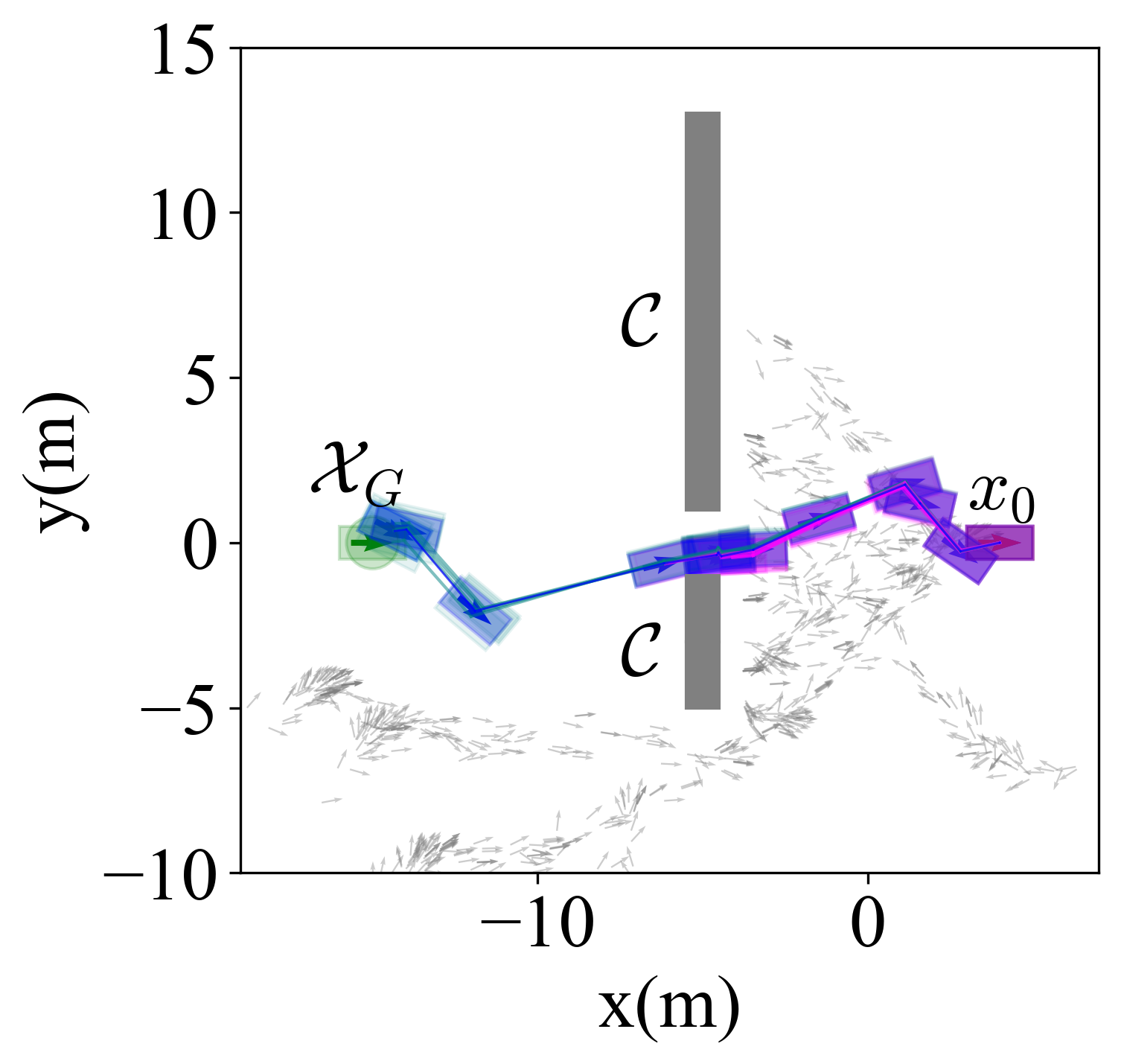}
        \end{subfigure}
        \begin{subfigure}[t]{0.176\textheight}
        \includegraphics[width=\textwidth,trim=72 0 0 0, clip]{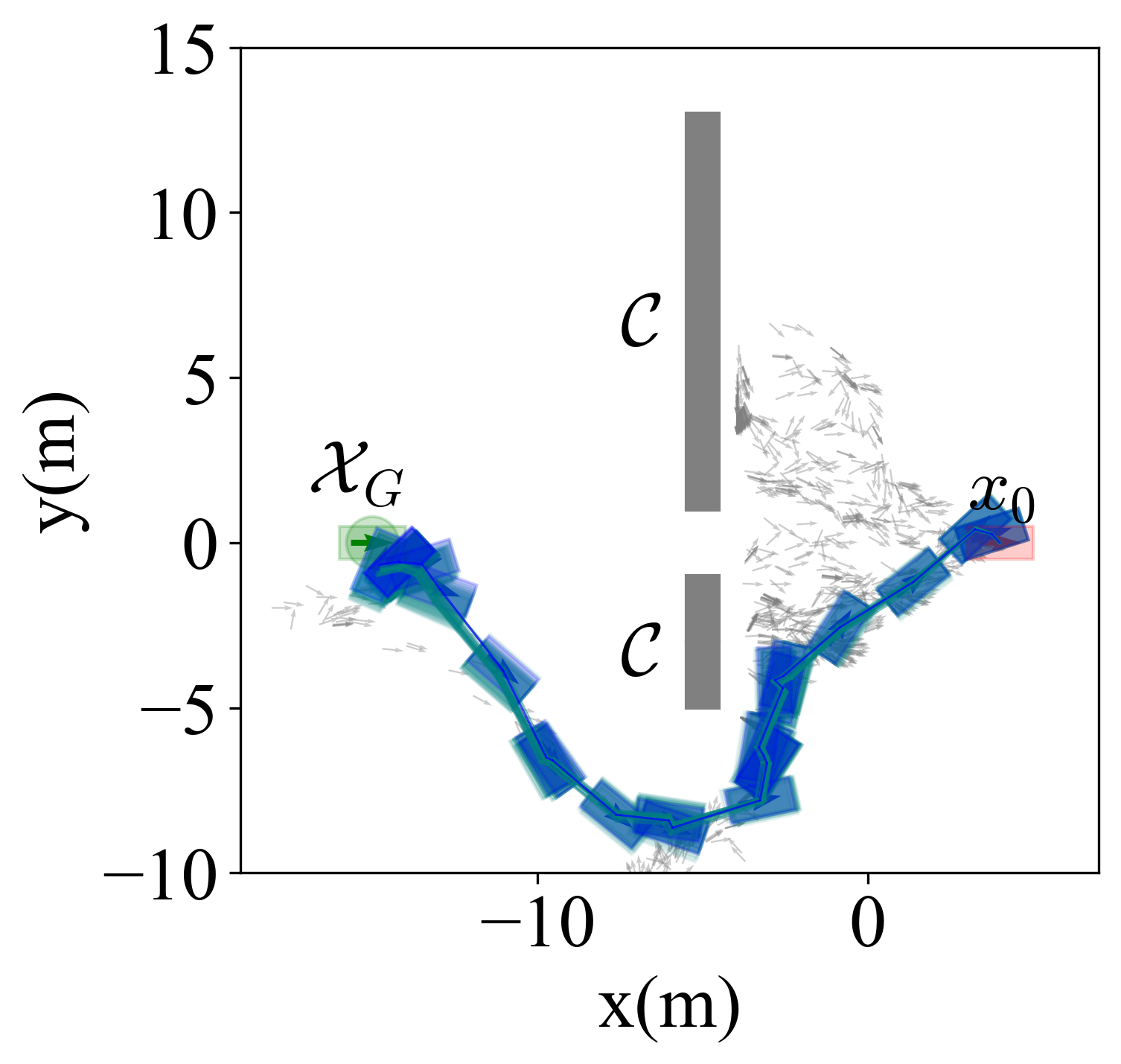}
        \end{subfigure}
    \caption{Planar pusher motion planning problem (left): the homotopy classes $\mathcal{H}_1,\mathcal{H}_3$ are robust, but $\mathcal{H}_2$ is not. \rrt selects $\mathcal{H}_2$ (middle) and \randuprrt chooses $\mathcal{H}_1$ (right). %
    We visualize the planned trajectory (blue) and 10 uncertainty realizations with $\|w_t\|\leq 0.5$ (teal: safe, magenta: collision with $\mathcal{C}$).}
    \label{fig:planar_pusher}
    \vspace{-5mm}
\end{figure}

\noindent
\textbf{Planar Pusher with Fixed Finger.} 
Consider the task of pushing a box %
to a desired pose using a fixed point contact. The contact force satisfies friction cone constraints and no ``pulling'' motion is allowed. A random bounded disturbance force is applied to the box. A bounded-size invariant set cannot be computed for this system because the control inputs cannot cancel all possible disturbance forces. To demonstrate the generality of our approach, we use PyBullet \cite{coumans2019pybullet} for parallelized black-box simulation of the system's dynamics (see \cite{Hogan2018ADA} for an analytical model).
%
% \vspace{-3mm}
The scenario and results are shown in Fig. \ref{fig:planar_pusher}. %
Three homotopy classes $\mathcal{H}_1$, $\mathcal{H}_2$, and $\mathcal{H}_3$ are available, where $\mathcal{H}_2$ is not robustly feasible due to tight obstacle clearance. Without considering uncertainty, \rrt explores $765$ nodes after $199.3$ seconds and returns a short but unsafe plan in $\mathcal{H}_2$: $52\%$ of the simulated rollouts from this plan result in collisions. 
In contrast, to account for uncertainty, \randuprrt explores $953$ nodes, eventually selecting a longer route in $\mathcal{H}_1$. \randuprrt planning only takes $2.24\times$ ($1.81\times$ longer per node) the runtime of \rrt. This results in a robust path with no infeasible rollouts. %

\begin{table}[!ht]
\vspace{-6mm}
\caption{Quadrotor planning statistics over $50$ runs. The means and standard deviations are shown for the node count. Validity is obtained by executing the motion plans with $10^4$ new rollouts. Each plan is valid only if all rollouts are safe and reach the goal. }
\centering    %
\vspace{2mm}
\begin{tabular}{|c|c|c|c|}
\hline
&\phantom{sss}Runtime\phantom{sss} & \phantom{ss}Node count\phantom{ss} & \phantom{sss}Validity\phantom{sss}\\     %
\hline
\phantom{ss}RRT, $0.3$-padding\phantom{ss}
& $1\times$ ($3.51$ seconds) & $2120\pm1020$ & $50\%$ \\
\hline
RRT, $0.5$-padding 
& $1.28\times$ & $2941\pm1911$ & $88\%$ \\ % 4.48 seconds
\hline
$\epsilon$-\randuprrt & $5.18\times$ & $2276\pm1700$ & $100\%$  \\ % 18.18 seconds
\hline
\end{tabular}
\label{tab:quadrotor_statistics}
% \vspace{-10mm}
\end{table}

\begin{figure}[!htb]
    \vspace{-5mm}
    \centering
        \begin{subfigure}[t]{0.45\textwidth}
        \includegraphics[width=\textwidth]{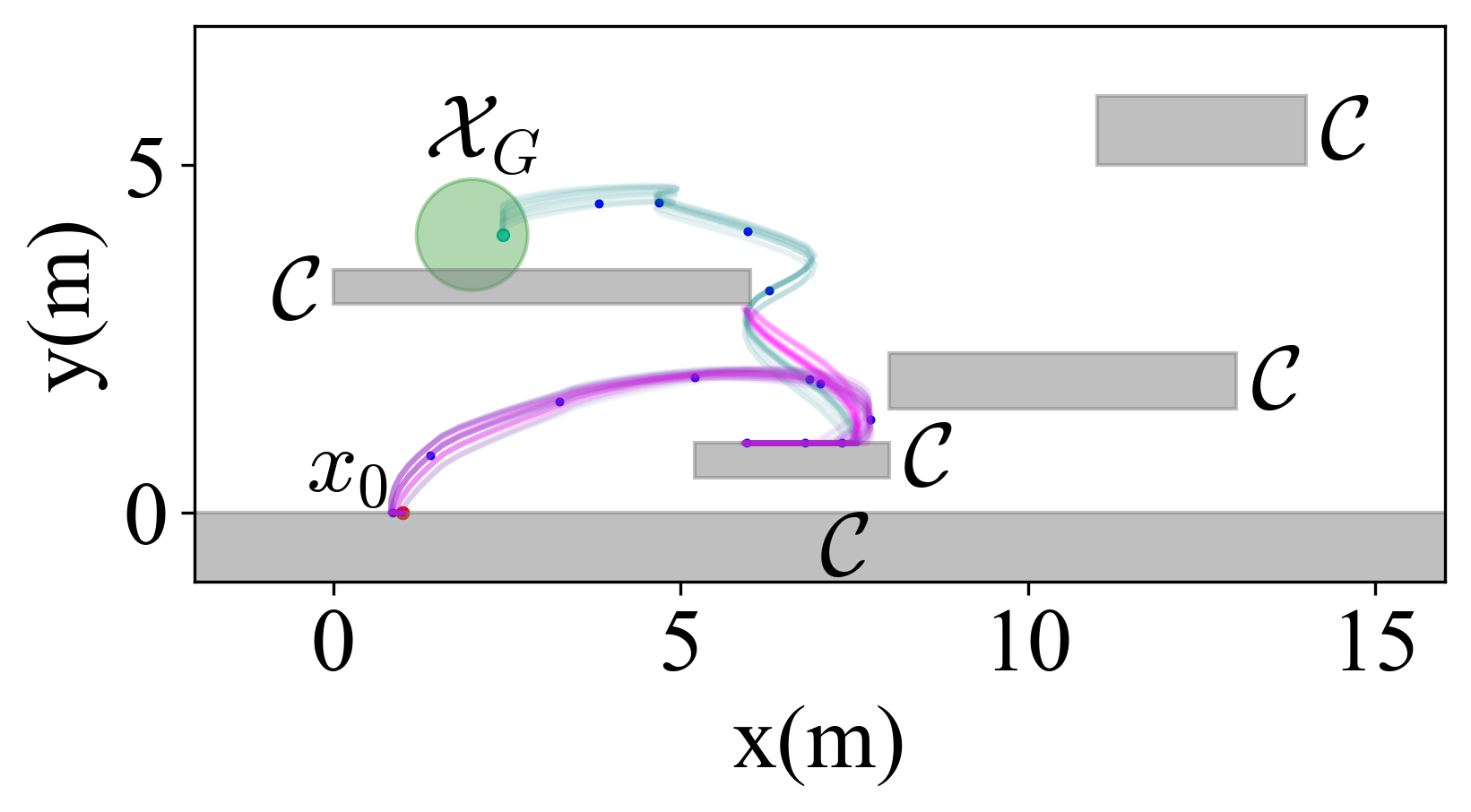}
        \end{subfigure}
        \begin{subfigure}[t]{0.45\textwidth}
        \includegraphics[width=\textwidth]{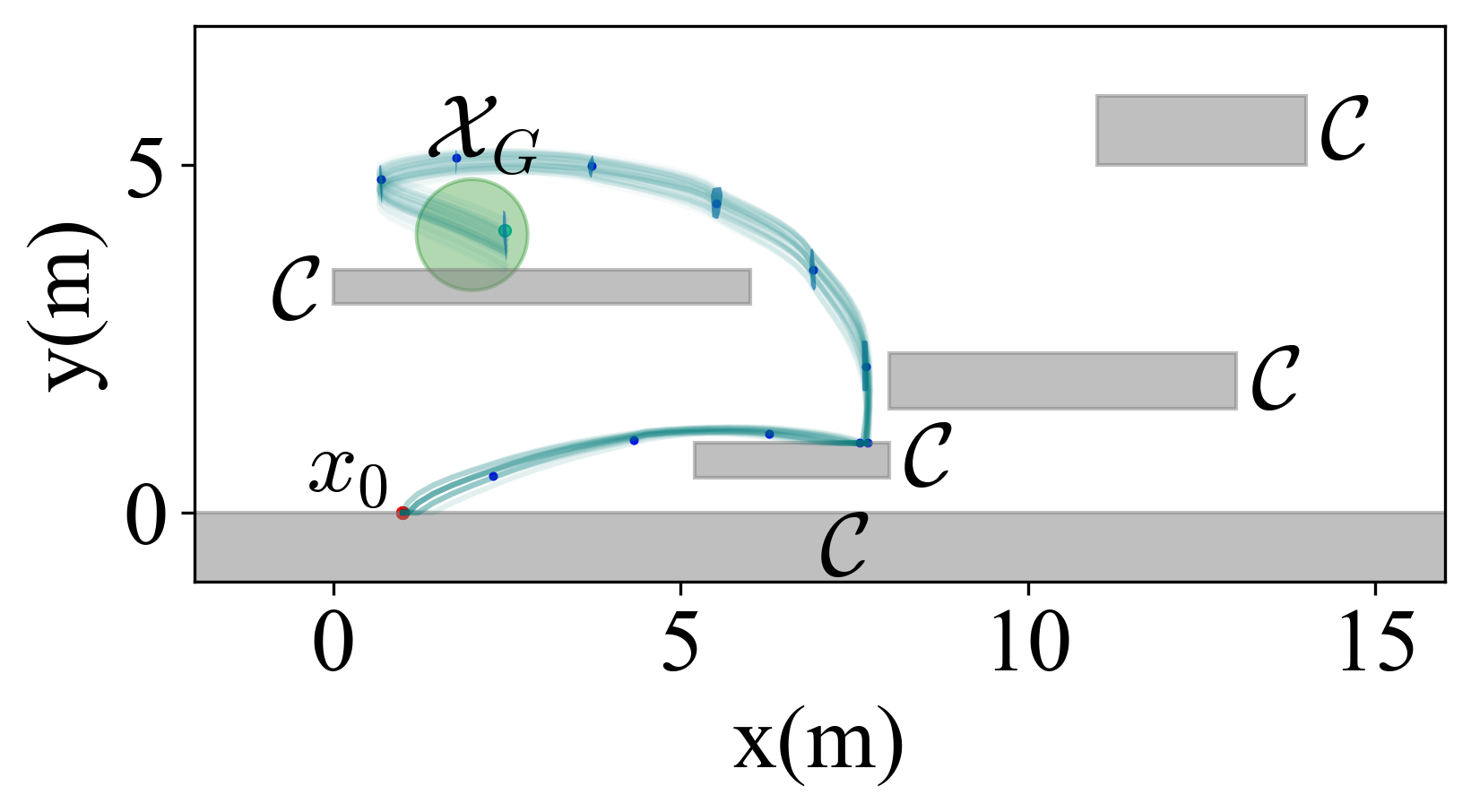}
        \end{subfigure}
    \vspace{-4mm}
    \caption{Hybrid jumping robot motion planning with \rrt(left, 7639 nodes) and \randuprrt(right, 9327 nodes).  %
    We overlay tree nodes (blue) with 50 rollouts (teal: safe. magenta: collision with $\mathcal{C}$). }
    \label{fig:hybrid_plan}
    \vspace{-9mm}
\end{figure}

\noindent
\textbf{Hybrid Jumping Robot.} 
\randuprrt is verified on a simulated jumping robot with bounded uncertain mass. %
The robot may be in two dynamic modes, \textit{contact} and \textit{flight}. The transition from \textit{contact} to \textit{flight} is subjected to a bounded random latency.
We present results in Fig. \ref{fig:hybrid_plan}. \rrt returns an unsafe trajectory in $16.6$ seconds: $56\%$ of simulated rollouts collide with obstacles. %
In contrast, \randuprrt requires larger computational time ($22.58\times$ that of \rrt) but returns a robustly feasible plan that solves the planning problem: $100\%$ of the rollouts avoid obstacles and safely reach the goal. 
This shows that \randuprrt is able to account for parametric and guard surface uncertainty.

\vspace{-3mm}
\section{Conclusion}
\label{sec:conclusion}
\vspace{-2mm}
\robustrrt provides a general and probabilistically complete sampling-based solution to robust motion planning. By jointly constructing a nominal and a robust tree using forward reachability analysis, \robustrrt accounts for model uncertainty with no additional conservatism and avoids computationally expensive set-distance computations. We propose a practical implementation, \randuprrt, that we validate on uncertain nonlinear and hybrid systems.

Extending the PC guarantees of \robustrrt to hybrid systems and\\ \randuprrt is of immediate interest, as is developing asymptotically optimal variants and extensions to uncertainty in state estimation \cite{bry2011rapidly,Aghamohammadi2014FIRMSF}.
We plan to validate \robustrrt on hardware systems where existing motion planners fall short, such as dexterous robotic manipulation under uncertainty.

\vspace{-4mm}
\bibliographystyle{splncs04}
\bibliography{ASL_papers,main}
\clearpage
\appendix

\section{Proof of Lemma \ref{lemma:lipschitz_continuity_all}: Reachable sets are Lipschitz}
\label{apdx:lipschitz_continuity_all_proof}
\begin{proof}
For ease of notations, we only consider uncertain parameters $\theta\in\Theta$; the conclusion follows similarly when considering disturbances $w_t\in\W$. Without loss of generality, we assume $t_1\leq t_2$.

We start by applying the triangle inequality (note that the Hausdorff distance is a metric over the space of nonempty compact sets), which allows decomposing \eqref{eq:reachable_set_deviation_full} as follows:
\begin{align}\label{eq:dh_Xt12:triangular}
  d_{\textrm{H}}(X_{t_1}^{u^1},X_{t_2}^{u^2}) 
    &\leq
     d_{\textrm{H}}(X_{t_1}^{u^1},X_{t_1}^{u^2}) + d_{\textrm{H}}(X_{t_1}^{u^2},X_{t_2}^{u^2}).
\end{align}
We bound the two terms as follows.

\textbf{First term:} First, we rewrite the left-hand side of the Hausdorff distance $d_{\textrm{H}}(X_{t_1}^{u^1},X_{t_2}^{u^2})$ in \eqref{eq:dh_Xt12:triangular} as follows:
    \begin{align}\label{eq:lhs_dh_reachsets}
    \nonumber
    &\sup_{x^1_t \in X_t^{u^1}(X_0^1)}
    \Inf_{x^2_t \in X_t^{u^2}(X_0^2)}
    \|x^1_t - x^2_t\|
    \\
    \nonumber
    &\hspace{5mm}=
    \sup_{\theta \in \Theta, x_0^1\in X_0^1}
    \Inf_{\tilde\theta \in \Theta, x_0^2\in X_0^2}
    \|{}^\theta x_t^{u^1}(x_0^1) 
    -
    {}^{\tilde\theta} x_t^{u^2}(x_0^2)\|
    \\
    &\hspace{5mm}\leq
    \sup_{\theta \in \Theta, x_0^1\in X_0^1}
    \Inf_{x_0^2\in X_0^2}
    \|{}^\theta x_t^{u^1}(x_0^1) 
    -
    {}^{\theta} x_t^{u^2}(x_0^2)\|,
    \end{align}
    since for any $\theta\in\Theta$ and any $x_0^1,x_0^2\in\R^n$,
    $$
    \Inf_{\tilde\theta\in\Theta}
    \|{}^\theta x_t^{u^1}(x_0^1)
    -
    {}^{\tilde\theta} x_t^{u^2}(x_0^2)\|
    \leq
    \|{}^\theta x_t^{u^1}(x_0^1) 
    -
    {}^\theta x_t^{u^2}(x_0^2)\|.
    $$
    Then, by Assumption \ref{assumption:lipschitz_bounded}, 
for any $\theta\in\Theta$, any two controls $u^1$ and $u^2$ and two initial conditions $x_0^1$ and $x_0^2$,
    \begin{equation}
    \label{eqn:lipschitz_intermediate_result}
    \|{}^\theta x_t^{u^1}(x_0^1)-{}^\theta x_t^{u^2}(x_0^2)\|\leq L_t^\theta(\|x_0^1-x_0^2\|+ \|u_1-u_2\|),
    \end{equation}
    where $t\in[0,t_2]$ and $L_t^\theta\geq 0$ is a bounded constant for any $\theta\in\Theta$ that stems from Assumption \ref{assumption:lipschitz_bounded} using a standard Lipschitz argument (see Lemma \ref{lemma:lipschitz_trajectory_final} in the Appendix \cite{Wu2022robustrrt}). 
    Let $L_t\triangleq \sup_{\theta\in\Theta}L_t^\theta$, which is bounded as $L_t^\theta$ and $\Theta$ are both bounded. Then,
    \begin{align*}
    \eqref{eq:lhs_dh_reachsets}\hspace{1mm}&\leq
    \sup_{\theta \in \Theta, x_0^1\in X_0^1}
    \Inf_{x_0^2\in X_0^2} L_t^\theta(\|x_0^1-x_0^2\|+ \|u_1-u_2\|)
    \\
    &\leq
    \left(\sup_{\theta \in \Theta} L_t^\theta\right)
    \left(\sup_{x_0^1\in X_0^1}
    \Inf_{x_0^2\in X_0^2} \|x_0^1-x_0^2\|+ \|u_1-u_2\|\right)
    \\
    &\leq
    L_t(d_{\textrm{H}}(X_0^1,X_0^2)+\|u_1-u_2\|)
    \\
    &\leq
    L_{t_2}(d_{\textrm{H}}(X_0^1,X_0^2)+\|u_1-u_2\|),
    \end{align*}
    where the last inequality holds since $L_t$ is monotonically increasing in $t$ (see Lemma \ref{lemma:lipschitz_trajectory_final}) and $t\leq t_2$. 
    By the definition of the Hausdorff distance which is symmetric, 
    we obtain $d_{\textrm{H}}(X_{t_1}^{u^1},X_{t_1}^{u^2})
    \leq 
     L_t(d_{\textrm{H}}(X_0^1,X_0^2)+\|u^1-u^2\|)$. 
     
     \textbf{Second term: } From Assumption \ref{assumption:lipschitz_bounded}, we have the inequality
\begin{align*}
    &\|x_{t_2}^{u}-x_{t_1}^{u}\| 
    \leq \left\| \int^{t_2}_{t_1}f_\theta(x^u_t, u) \d t \right\|
    \leq \lambda(t_2-t_1),
\end{align*}
where the last inequality is obtained with the Cauchy-Schwartz Inequality.
\begin{align*}
    \left\| \int^{t_2}_{t_1}f_\theta(x^u_t, u) \d t\right\|^2
    &=
    \left\| \int_{\mathbb{R}} \mathbbm{1}_{[t_1,t_2]}(t)^2f_\theta(x^u_t, u) \d t\right\|^2
    \\
    &\hspace{-2cm}\leq
     \int_{\mathbb{R}} \left\|\mathbbm{1}_{[t_1,t_2]}(t)f_\theta(x^u_t, u)\right\|^2 \d t
    \int_{\mathbb{R}}|\mathbbm{1}_{[t_1,t_2]}(t)|^2 \d t
    \\
    &\hspace{-2cm}=
     \int^{t_2}_{t_1}\left\|f_\theta(x^u_t, u)\right\|^2 \d t
     \cdot
    (t_2-t_1)
    \\
    &\hspace{-2cm}\leq
     \int^{t_2}_{t_1}\lambda^2 \d t
     \cdot
    (t_2-t_1)
    \\
    &\hspace{-2cm}=
    \lambda^2
    (t_2-t_1)^2.
\end{align*}
\textbf{Combining the two terms: } Using the last two inequalities, we define $L=\max(L_{t_2}, \lambda)$ and conclude using \eqref{eq:dh_Xt12:triangular}.
\end{proof}

\section{Lipschitz trajectories}
\begin{lemma}
\label{lemma:lipschitz_trajectory_final}
Let  $x_0^1,x_0^2\in\R^n$ be two initial states, 
$u^1,u^2$ be two control inputs of durations $t_1,t_2\geq 0$, respectively, such that $u_t^i=u_i$ for all $t\in[0,t_i)$ with $i=1,2$, and define
$$
x^i(t)=x_0^i+\int_{0}^{t} f_\theta(x^i(\tau), u^i(\tau),w(\tau))\textrm{d}\tau,
$$
where $i=1,2$, 
$t\in[0,t_i]$, 
 $\theta\in\Theta$, and 
 $w\in\W$.%
 
 Then, under Assumption \ref{assumption:lipschitz_bounded}, for any $t\in[0,\min(t_1,t_2)]$,
 \begin{align*}
\|x^1_t-x^2_t\|
&\leq 
L_t\left(
\|x_0^1{-}x_0^2\|
+
\|u_1-u_2\|
\right),
\end{align*}
where $L:[0,\min(t_1,t_2)]\rightarrow \R_{\geq 0}$ is continuous and monotonically increasing.
\end{lemma}
\begin{proof}

 As the parameters and disturbances are fixed, we denote $\bar{f}(x,u)=f_\theta(x,u,w)$ for conciseness. We denote $x^i(t)=x^i_t$. 
To prove \eqref{eqn:lipschitz_intermediate_result}, we proceed as follows:

\begin{align*}
\frac{\|x^1_t-x^2_t\|^2}{2}
&\leq 
\|x_0^1-x_0^2\|^2+\left\|\int_{0}^{t} (\bar{f}(x^1_\tau, u^1_\tau){-}\bar{f}(x^2_\tau, u^2_\tau))\textrm{d}\tau\right\|^2
\\
&\hspace{-1.6cm}= 
\|x_0^1{-}x_0^2\|^2
{+}
\left\|\int_{\R} 
\mathbf{1}_{[0,t]}(\tau)(\bar{f}(x^1_\tau, u^1_\tau)-\bar{f}(x^2_\tau, u^2_\tau))\textrm{d}\tau\right\|^2
\\
&\hspace{-1.6cm}\leq 
\|x_0^1{-}x_0^2\|^2
{+}
\int_{\R}\hspace{-1mm} \mathbf{1}_{[0,t]}(\tau)^2\textrm{d}\tau
\int_{0}^{t}\hspace{-1mm}
\|\bar{f}(x^1_\tau, u^1_\tau){-}\bar{f}(x^2_\tau, u^2_\tau)\|^2\textrm{d}\tau
\\ %
&\hspace{-1.6cm}\leq 
\|x_0^1{-}x_0^2\|^2
{+}
2tK^2
\int_{0}^{t} (\|x^1_\tau-x^2_\tau\|^2 + 
\|u^1_\tau-u^2_\tau\|^2)\textrm{d}\tau,
\end{align*}
where we used the triangular and Cauchy-Schwarz inequalities, Assumption \ref{assumption:lipschitz_bounded}, and $(a+b)^2 \leq 2 (a^2 + b^2)$. 
To obtain \eqref{eqn:lipschitz_intermediate_result}, we apply 
Gr\"onwall's inequality so that
\begin{align*}
\|x^1_t-x^2_t\|^2
&\leq 
2e^{
K^2t^2
}(\|x_0^1{-}x_0^2\|^2{+}2tK^2\int_{0}^{t}
\|u^1_\tau-u^2_\tau\|^2
\textrm{d}\tau)
\\
&=
2e^{
K^2t^2
}(\|x_0^1{-}x_0^2\|^2{+}2t^2K^2
\|u_1-u_2\|^2
)
,
\end{align*}
where the second equality holds since the control inputs are piecewise constant. 
Defining $A_t=\max(1,2t^2K^2)$, we obtain  
$$
\|x^1_t-x^2_t\|^2
\leq 
2e^{
K^2t^2
}A_t(\|x_0^1{-}x_0^2\|^2
+
\|u_1-u_2\|^2)
.
$$
Then, since $\sqrt{a+b}\leq \sqrt{a}+\sqrt{b}$, we obtain \begin{align*}
\|x^1_t-x^2_t\|
&\leq 
L_t\left(
\|x_0^1{-}x_0^2\|
+
\|u_1-u_2\|
\right),
\end{align*}
where $L_t=\sqrt{2A_t}e^{Kt}$. This concludes the proof. 
\end{proof}

\section{Hybrid Adaptations of \robustrrt}
\label{apdx:hybrid_adaptations}
\robustrrt can be generalized to mode-explicit planning in hybrid systems. At each extension, the desired dynamic mode is explicitly sampled, and all resulting states must belong to the desired dynamic mode for successful tree extension.  We empirically demonstrate the algorithm's ability to account for uncertain guard surface locations in Section \ref{sec:experiments}. We leave the theoretical properties of \robustrrt in hybrid systems to future work.

To adapt \robustrrt (Algorithm \ref{alg:robust_rrt}) to hybrid systems, we replace Line \ref{algline:rrt_node_sampling:control} of \ref{alg:robust_rrt} with Algorithm \ref{alg:sample_control_hybrid} and perform tree extension with Algorithm \ref{alg:extend_hybrid}.

\begin{figure*}[b]
\centering
\begin{minipage}{0.495\linewidth}
\begin{algorithm}[H]
\caption{\texttt{sample_control_hyb()} }
\begin{algorithmic}[1]
\Require{$\U$, $\tau_{\textrm{max}}, X_{s}$}
\Ensure{$\bar{u}, \tau, \sigma$}
\State{Sample $\bar{u}\sim \textrm{Unif}(\U)$}
\State{Sample $\tau\sim \textrm{Unif}([0,\tau_{\textrm{max}}])$}
\State{Sample $\sigma\sim \textrm{Unif}(\Sigma(X_s))$}
\Comment{Sample mode from all possible resulting modes from $X_s$}
\State{\textbf{return} $(\bar{u},\tau, \sigma)$}
\end{algorithmic}
\label{alg:sample_control_hybrid}
\end{algorithm}
\end{minipage}
\hspace{0.008\linewidth}
\begin{minipage}{0.48\linewidth}
\centering
\begin{algorithm}[H]
\caption{\texttt{extend_hybrid()} }
\begin{algorithmic}[1]
\Require{$X_s, \hat{x}_s, \bar{u}, \tau, \sigma_s$}
\Ensure{$X_{new}, \hat{x}_{new}$}
\State{$\hat{x}_{new} \gets \hat{x}_s + \int_0^\tau f_{\hat{\theta}}(x_{\tau'}, \bar{u}, \hat{w})d\tau'$}
\State{$\hat{\sigma}_{new} \gets r(\hat{x}_s, \bar{u}, \hat{w}, \tau)$}
\If{$\hat{\sigma}_{new} \neq \sigma_s$}
    \State{\textbf{return} $\emptyset$}
\EndIf
\State{$X_{new}{\gets}\texttt{compute_reach_set}(X_s, \bar{u}, \tau)$}
\If{$\exists \sigma_{new} \neq \sigma_s, \forall (x_{new}, \sigma_{new}) \in X_{new}$}
\State{\textbf{return} $\emptyset$}
\EndIf
\State{\textbf{return} $(X_{new}, \hat{x}_{new})$}
\end{algorithmic}
\label{alg:extend_hybrid}
\end{algorithm}
\end{minipage}
\end{figure*}

\section{Implementation Details}
\label{apdx:implementation_details}
All experiments are performed on a laptop computer with an Intel Core i7-7820HQ CPU (8 threads) and 16GB RAM.
For simplicity and computational speed, the nominal states are computed as the arithmetic mean of all \randup particles. This choice does not affect the theoretical guarantees. $\mathcal{B}_{\zeta}$ can still be constructed for each nominal state, and one can show that the lower bound on \eqref{eqn:prob_successful_extension_indep} still holds.
\subsection{Nonlinear Quadrotor Setup}
\label{apdx:quadrotor}
Denoting
$[p_x,p_y,v_x,v_y]^\top$ for the state of the system 
and 
$[\tan(\theta), \tan(\phi)]^\top$ for the control input,  
where 
$\theta$ is the robot's pitch, $\phi$ is its roll ($\phi$), and $\tau-g$ is its gravity-compensated thrust, 
the system's 
nonlinear dynamics are given by %
\begin{equation}\label{eq:quad:dynamics}
   \dot{x}_t=
   Ax_t
   +
   Bu_t
   +
   d(x_t),
   \quad t\in\R,
\end{equation}
with 
$$
A=\begin{bmatrix}
   0_{2\times 2} & I_{2\times 2}
   \\
   0_{2\times 2} & 0_{2\times 2}
   \end{bmatrix},
   \quad 
   B=
   \begin{bmatrix}
   0_{1\times 2} & g & 0
   \\
   0_{1\times 2} & 0 & -g
   \end{bmatrix}^\top,
   \quad 
d(x_t) = \begin{bmatrix}
0\\
0\\
-\alpha_x v_x |v_x|\\
-\alpha_y v_y |v_y|
\\
\end{bmatrix}
$$
The drag model $d(x_t)$ is adapted from \cite{RichardsAzizanEtAl2021}, $\alpha = [\alpha_x, \alpha_y]^T \in \mathbb{R}^2$ is the uncertain drag coefficient.
Using a zero-order hold over $\Delta t$ on the control input, i.e., $u_t=u_{t_k}$ for all $t\in[t_k,t_{k}+T)$, $k\in\mathbb{N}$, we discretize the %
dynamics in \eqref{eq:quad:dynamics} with an approximate two-steps forward Euler scheme  
\begin{align}\label{eq:quad:dynamics:dt}
x_{t+\Delta t} 
&= 
(I{+}A(\scalebox{0.9}{$\frac{\Delta t}{2}$}))^2x_t + (\Delta t{\cdot}I{+}A(\scalebox{0.9}{$\frac{\Delta t}{2}$})^2)B u_{t}
+
Td(x_t).
\end{align}
We simulate these discrete-time dynamics in experiments. 
To reduce uncertainty, we optimize over open-loop control inputs $\nu_t\in\R^m$ with associated nominal trajectory $\mu_t$ where $\mu_{t+\Delta t}=\eqref{eq:quad:dynamics:dt}(\mu_t,\nu_t)$ and apply the linear feedback control law $u_t=\nu_t + K(x_t-\mu_t)$ to the true system in \eqref{eq:quad:dynamics:dt}.  This standard approach enables considering the reduction in uncertainty due to feedback while only searching for the nominal control inputs $\nu_t$. %

Starting from the origin $x_0=0$, the problem consists of reaching a goal region $\X_G=\{(p,v)\in\R^4: \|p-(10, 0)\|\leq 0.7\}$ while avoiding a set of spherical obstacles of radius $2.3$. 

For the \randuprrt quadrotor experiments, we use 100 \randup particles with a sampled fixed value of $\alpha$. We used $(\epsilon=0.3)$-padding, as computing conservative reachable set estimates with \randup requires an additional padding step, see \cite{LewJansonEtAl2022} for further details. To implement this step, we simply pad all obstacles by $\epsilon$ and shrink the goal region inwards by $\epsilon$.

\subsection{Planar Pusher Setup}
\label{apdx:pusher}
The state space for the planner is in $\mathbb{R}^6$. During planning, \texttt{Robust-RRT} samples a desired state $x_{d}\in \mathbb{R}^6$, and a predefined PD feedback controller steers the box toward $x_d$. The environment shown in Fig. \ref{fig:planar_pusher} has the following dimensions. The object has size $2m \times 1m$. The gap between the obstacles is $2m$. The starting state is at $(4,0,0,0,0,0)$. The goal state is at $(-15,0,0,0,0,0)$.

The motion of the object is affected by the following attributes: ground friction with coefficient $\mu_g = 1.2$, the pushing force control input $u$, and a random bounded disturbance $w \in \mathcal{W}=\{w\mid\,|w| \leq w_{max}\}\subset \mathbb{R}^2$. $u$ is applied at a fixed location on the long axis of the object. The control input $u$ is subjected to the friction cone constraints $u \in \mathcal{U} \subset \mathbb{R}^2$, where $\mathcal{U}$ is the friction cone in Equation \eqref{eqn:friction_cone}. %
Here, $f_N$ is the normal force at the contact point with the positive direction pointing into the object, and $f_T$ is the tangent force at the contact point. The friction coefficient is set to $\mu = 0.5$ and is known to the planner. Note that there is a separate friction coefficient $\mu_g$ between the object and the ground that is fixed but unknown.
\begin{equation}
    \mathcal{U} = \left\{\left[\begin{smallmatrix} f_N \\ f_T \end{smallmatrix}\right] \; \middle| \; f_N \geq 0,\; |f_T| \leq \mu f_N\right\}.
    \label{eqn:friction_cone}
\end{equation}
Disturbances $w_t$ are applied with a zero-order hold at the center of mass of the object, which therefore do not induce any torque. The value $w_t$ is resampled from $\mathcal{W}$ every $1/3$ second in simulation time.
All \randuprrt planning for the planar pusher is performed with $16$ \randup particles with no $\epsilon$-padding. The simulation is parallelized across 8 threads.

\subsection{Hybrid Jumping Robot Setup}
\label{apdx:hybrid_planning}
The system is simulated under time step size $\tau=0.03$s. PD local controllers are used to stabilize the horizontal position $x$, and each local controller is held for $7\tau$. The \textit{contact} dynamics are given below:
\begin{align*}
\begin{split}
    x_{k+1} &= x_k + \dot{x}_k\tau,\; \dot{x}_{k+1} = \dot{x}_k + u\tau, \\
    y_{k+1} &= y_{k},\; \dot{y}_{k+1} = 0.
\end{split}
\end{align*}
The \textit{flight} mode dynamics have modified dynamics on $y$:
\begin{align*}
\begin{split}
    y_{k+1} &= y_{k}+\dot{y}_k\tau, \; \dot{y}_{k+1} = \dot{y}_k-g\tau.
\end{split}
\end{align*}
In total, 100 \randup particles with no $\epsilon$-padding are used for \randuprrt experiments in the hybrid jumping robot.
\end{document}

%% file: macros.tex
\DeclareMathOperator*{\argmin}{arg\,min}
\newtheorem{assumption}{Assumption}
\theoremstyle{definition}

\usepackage{xspace}
\newcommand{\randuprrt}{\textsc{RandUP-RRT}\xspace}
\newcommand{\robustrrt}{\textsc{Robust-RRT}\xspace}
\newcommand{\rrt}{\textsc{RRT}\xspace}
\newcommand{\randup}{\textsc{RandUP}\xspace}

\newcommand{\R}{\mathbb{R}}

\newcommand{\N}{\mathbb{N}}

\newcommand{\Obs}{\mathcal{C}}
\newcommand{\Prob}{\mathbb{P}}

\newcommand{\W}{\mathcal{W}}
\newcommand{\U}{\mathcal{U}}
\newcommand{\X}{\mathcal{X}}

\def\d{\textrm{d}}
\def\Inf{\operatornamewithlimits{inf\vphantom{p}}}

%% file: main.bbl
\newcommand{\noopsort}[1]{} \newcommand{\printfirst}[2]{#1}
  \newcommand{\singleletter}[1]{#1} \newcommand{\switchargs}[2]{#2#1}
\begin{thebibliography}{10}
\providecommand{\url}[1]{\texttt{#1}}
\providecommand{\urlprefix}{URL }
\providecommand{\doi}[1]{https://doi.org/#1}

\bibitem{abbasi2011improved}
Abbasi-Yadkori, Y., P{\'a}l, D., Szepesv{\'a}ri, C.: Improved algorithms for
  linear stochastic bandits. In: {Conf.\ on Neural Information Processing
  Systems} (2011)

\bibitem{Aghamohammadi2014FIRMSF}
Agha-mohammadi, A., Chakravorty, S., Amato, N.: {FIRM}: Sampling-based feedback
  motion-planning under motion uncertainty and imperfect measurements. The Int.
  Journal of Robotics Research  \textbf{33}(2),  268--304 (2014)

\bibitem{Althoff2021}
Althoff, M., Frehse, G., Girard, A.: Set propagation techniques for
  reachability analysis. {Annual Review of Control, Robotics, and Autonomous
  Systems}  \textbf{4}(1),  369--395 (2021)

\bibitem{bansal2017hamilton}
Bansal, S., Chen, M., Herbert, S., Tomlin, C.J.: {Hamilton-Jacobi}
  reachability: A brief overview and recent advances. In: Conference on
  Decision and Control (2017)

\bibitem{Hogan2018ADA}
Bauz{\'a}, M., Hogan, F.R., Rodr{\'i}guez, A.: A data-efficient approach to
  precise and controlled pushing. In: Conference on Robot Learning (2018)

\bibitem{bry2011rapidly}
Bry, A., Roy, N.: Rapidly-exploring random belief trees for motion planning
  under uncertainty. In: Int. Conf. on Robotics and Automation. pp. 723--730.
  IEEE (2011)

\bibitem{chow1940systeme}
Chow, W.L.: {\"U}ber systeme von liearren partiellen differentialgleichungen
  erster ordnung. Mathematische Annalen  \textbf{117}(1),  98--105 (1940)

\bibitem{coumans2019pybullet}
Coumans, E., Bai, Y.: Pybullet, a python module for physics simulation for
  games, robotics and machine learning. \url{http://pybullet.org} (2016--2019)

\bibitem{danielson2020robust}
{Danielson}, C., {Berntorp}, K., {Weiss}, A., {Cairano}, S.D.: Robust motion
  planning for uncertain systems with disturbances using the invariant-set
  motion planner. IEEE Transactions on Automatic Control  \textbf{65}(10),
  4456--4463 (2020)

\bibitem{herbert2017fastrack}
Herbert, S.L., Chen, M., Han, S., Bansal, S., Fisac, J.F., Tomlin, C.J.:
  {FaSTrack}: A modular framework for fast and guaranteed safe motion planning.
  In: Conference on Decision and Control (2017)

\bibitem{jaillet2011eg}
Jaillet, L., Hoffman, J., Van~den Berg, J., Abbeel, P., Porta, J.M., Goldberg,
  K.: {EG-RRT}: Environment-guided random trees for kinodynamic motion planning
  with uncertainty and obstacles. In: Int. Conf. on Intelligent Robots and
  Systems (2011)

\bibitem{kalise2020robust}
Kalise, D., Kundu, S., Kunisch, K.: Robust feedback control of nonlinear pdes
  by numerical approximation of high-dimensional {Hamilton--Jacobi--Isaacs}
  equations. SIAM Journal on Applied Dynamical Systems  \textbf{19}(2),
  1496--1524 (2020)

\bibitem{karaman2011sampling}
Karaman, S., Frazzoli, E.: Sampling-based algorithms for optimal motion
  planning. The Int. Journal of robotics research  \textbf{30}(7),  846--894
  (2011)

\bibitem{kleinbort2018}
{Kleinbort}, M., {Solovey}, K., {Littlefield}, Z., {Bekris}, K.E., {Halperin},
  D.: Probabilistic completeness of {RRT} for geometric and kinodynamic
  planning with forward propagation. Robotics and Automation Letters
  \textbf{4}(2) (2019)

\bibitem{lathrop2021distributionally}
Lathrop, P., Boardman, B., Mart{\'\i}nez, S.: Distributionally safe path
  planning: Wasserstein safe {RRT}. Robotics and Automation Letters
  \textbf{7}(1),  430--437 (2021)

\bibitem{lavalle_2001}
LaValle, S.M., James J.~Kuffner, J.: Randomized kinodynamic planning. The Int.
  Journal of Robotics Research  \textbf{20}(5),  378--400 (2001)

\bibitem{LewJansonEtAl2022}
Lew, T., Janson, L., Bonalli, R., Pavone, M.: A simple and efficient
  sampling-based algorithm for general reachability analysis. In: Learning for
  Dynamics \& Control Conference (2022)

\bibitem{lew2020samplingbased}
Lew, T., Pavone, M.: Sampling-based reachability analysis: A random set theory
  approach with adversarial sampling. In: Conference on Robot Learning (2020)

\bibitem{li2016asymptotic}
Li, Y., Littlefield, Z., Bekris, K.E.: Asymptotically optimal sampling-based
  kinodynamic planning. The Int. Journal of Robotics Research  \textbf{35}(5),
  528--564 (2016)

\bibitem{lindemann2021robust}
Lindemann, L., Cleaveland, M., Kantaros, Y., Pappas, G.J.: Robust motion
  planning in the presence of estimation uncertainty. arXiv preprint 2108.11983
   (2021)

\bibitem{liu2014incremental}
Liu, W., Ang, M.H.: Incremental sampling-based algorithm for risk-aware
  planning under motion uncertainty. In: Int. Conf. on Robotics and Automation
  (2014)

\bibitem{Luders2011ProbabilisticFF}
Luders, B., How, J.: Probabilistic feasibility for nonlinear systems with
  non-gaussian uncertainty using {RRT}. In: AIAA Aerospace Conference (2011)

\bibitem{luders2010chance}
Luders, B., Kothari, M., How, J.: Chance constrained {RRT} for probabilistic
  robustness to environmental uncertainty. AIAA Guidance, Navigation, and
  Control Conference  (2010)

\bibitem{luders2014optimizing}
Luders, B.D., How, J.P.: An optimizing sampling-based motion planner with
  guaranteed robustness to bounded uncertainty. In: American Control Conference
  (2014)

\bibitem{majumdar2016}
Majumdar, A., Tedrake, R.: Funnel libraries for real-time robust feedback
  motion planning. The Int. Journal of Robotics Research  \textbf{36}(8) (2016)

\bibitem{melchior2007particle}
{Melchior}, N.A., {Simmons}, R.: Particle {RRT} for path planning with
  uncertainty. In: Int. Conf. on Robotics and Automation (2007)

\bibitem{panchea2017extended}
Panchea, A.M., Chapoutot, A., Filliat, D.: Extended reliable robust motion
  planners. In: Conference on Decision and Control (2017)

\bibitem{Papadopoulos2014}
Papadopoulos, G., Kurniawati, H., Patrikalakis, N.M.: Analysis of
  asymptotically optimal sampling-based motion planning algorithms for
  lipschitz continuous dynamical systems. Annual Review of Control, Robotics,
  and Autonomous Systems  \textbf{3},  295--318 (2014)

\bibitem{pepy2009reliable}
Pepy, R., Kieffer, M., Walter, E.: Reliable robust path planning. Int. Journal
  of Applied Mathematics and Computer Science  \textbf{19}(3),  413--424 (2009)

\bibitem{RichardsAzizanEtAl2021}
Richards, S.M., Azizan, N., Slotine, J.J.E., Pavone, M.:
  Adaptive-control-oriented meta-learning for nonlinear systems. In: {Robotics:
  Science and Systems} (2021)

\bibitem{sadraddini2019sampling}
Sadraddini, S., Tedrake, R.: Sampling-based polytopic trees for approximate
  optimal control of piecewise affine systems. In: Int. Conf. on Robotics and
  Automation (2019)

\bibitem{shkolnik2009reachability}
Shkolnik, A., Walter, M., Tedrake, R.: Reachability-guided sampling for
  planning under differential constraints. In: Int. Conf. on Robotics and
  Automation (2009)

\bibitem{SinghChenEtAl2018}
Singh, S., Chen, M., Herbert, S.L., Tomlin, C.J., Pavone, M.: Robust tracking
  with model mismatch for fast and safe planning: an {SOS} optimization
  approach. In: {Workshop on Algorithmic Foundations of Robotics} (2018)

\bibitem{summers2018}
{Summers}, T.: Distributionally robust sampling-based motion planning under
  uncertainty. In: Int. Conf. on Intelligent Robots and Systems (2018)

\bibitem{tedrake2010lqr}
Tedrake, R., Manchester, I.R., Tobenkin, M., Roberts, J.W.: {LQR}-trees:
  Feedback motion planning via sums-of-squares verification. The Int. Journal
  of Robotics Research  \textbf{29}(8),  1038--1052 (2010)

\bibitem{verginis2021kdf}
Verginis, C.K., Dimarogonas, D.V., Kavraki, L.E.: {KDF}: Kinodynamic motion
  planning via geometric sampling-based algorithms and funnel control. arXiv
  preprint 2104.11917  (2021)

\bibitem{wang2020moment}
Wang, A., Jasour, A., Williams, B.: Moment state dynamical systems for
  nonlinear chance-constrained motion planning. arXiv preprint 2003.10379
  (2020)

\bibitem{wu2020r3t}
{Wu}, A., {Sadraddini}, S., {Tedrake}, R.: {R3T}: Rapidly-exploring random
  reachable set tree for optimal kinodynamic planning of nonlinear hybrid
  systems. In: Int. Conf. on Robotics and Automation (2020)

\bibitem{Wu2022robustrrt}
Wu, A., Lew, T., Solovey, K., Schmerling, E., Pavone, M.: Robust-rrt:
  Probabilistically-complete motion planning for uncertain nonlinear systems.
  arXiv preprint 2205.07728  (2022)

\end{thebibliography}
